\documentclass[letterpaper]{article} 
\usepackage{aaai23}  
\usepackage{times}  
\usepackage{helvet}  
\usepackage{courier}  
\usepackage[hyphens]{url}  
\usepackage{graphicx} 
\usepackage{natbib}  
\usepackage{caption} 
\usepackage{algorithm}
\usepackage{algorithmic}

\usepackage{subfig}
\usepackage{amsmath}
\usepackage{amssymb}
\DeclareMathOperator*{\Max}{max}

\DeclareMathOperator*{\argmax}{arg\,max}

\usepackage{bm}
\usepackage{pifont}
\newcommand{\cmark}{\ding{51}}%
\newcommand{\xmark}{\ding{55}}%
\newcommand{\ud}{\mathrm{d}}
\usepackage{multirow}
\usepackage{booktabs}
\usepackage{placeins}
\usepackage{newfloat}
\usepackage{listings}
\DeclareCaptionStyle{ruled}{labelfont=normalfont,labelsep=colon,strut=off} 
\floatstyle{ruled}
\newfloat{listing}{tb}{lst}{}
\floatname{listing}{Listing}
\nocopyright 

\title{Dynamic Representation Learning with Temporal Point Processes for Higher-Order Interaction Forecasting}

\author {
    Tony Gracious,
    Ambedkar Dukkipati
}
\affiliations {
    Indian Institute of Science, Bangalore - 560012, INDIA \\
    \{tonygracious, ambedkar\}@iisc.ac.in
}

\usepackage{bibentry}

\begin{document}

\maketitle
\begin{abstract}
The explosion of digital information and the growing involvement of people in social networks led to enormous research activity to develop methods that can extract meaningful information from interaction data. Commonly, interactions are represented by edges in a network or a graph, which implicitly assumes that the interactions are pairwise and static. However, real-world interactions deviate from these assumptions: (i) interactions can be multi-way, involving more than two nodes or individuals (e.g., family relationships, protein interactions), and (ii) interactions can change over a period of time (e.g., change of opinions and friendship status). While pairwise interactions have been studied in a dynamic network setting and multi-way interactions have been studied using hypergraphs in static networks, there exists no method, at present, that can predict multi-way interactions or hyperedges in dynamic settings. Existing related methods cannot answer temporal queries like what type of interaction will occur next and when it will occur. This paper proposes a temporal point process model for hyperedge prediction to address these problems. Our proposed model uses dynamic representation learning techniques for nodes in a neural point process framework to forecast hyperedges. We present several experimental results and set benchmark results. As far as our knowledge, this is the first work that uses the temporal point process to forecast hyperedges in dynamic networks. 
\end{abstract}
\section{Introduction} 
\label{sec: introduction}
Learning from temporal interactions between entities to extract meaningful information and knowledge is of paramount importance. For example, learning how a person interacts on social media can provide knowledge about that person's preferences, and it can help in recommending items to that person. Similarly, an e-commerce website can better understand the users' needs if it efficiently extracts knowledge from users' consumption history. Previously these problems have been studied using representation or embedding learning in dynamic networks where interactions were modeled as instantaneous links or edges between two nodes formed at the time of interaction~\citep{Nguyen:EtAL:2018:Continuous-time_dynamic_network_embeddings,Kumar:EtAL:2019:Predicting_Dynamic_Embedding_Trajectory_in_Temporal_Interaction_Networks}. 
For this, Temporal Point Processes (TPP)~\citep{Daley:EtAL:An_introduction_to_the_theory_of_point_processes.} have been introduced for modeling edge formation in dynamic networks. TPPs are stochastic processes that model localized events in time, and the events can be of multiple types. To model dynamic networks, one represents each edge as an event type, and a probability distribution is defined over the time of its formation. Here, the probability distribution is parameterized using an intensity function based on representations of nodes. These node representations are functions of time and past interaction events. The parameters of these functions are learned by minimizing the negative log-likelihood of the interactions in the training data.

\begin{figure*}[t]
	\centering
	\subfloat[][Hyperedges at time $t$]{\includegraphics[width=0.37\textwidth]{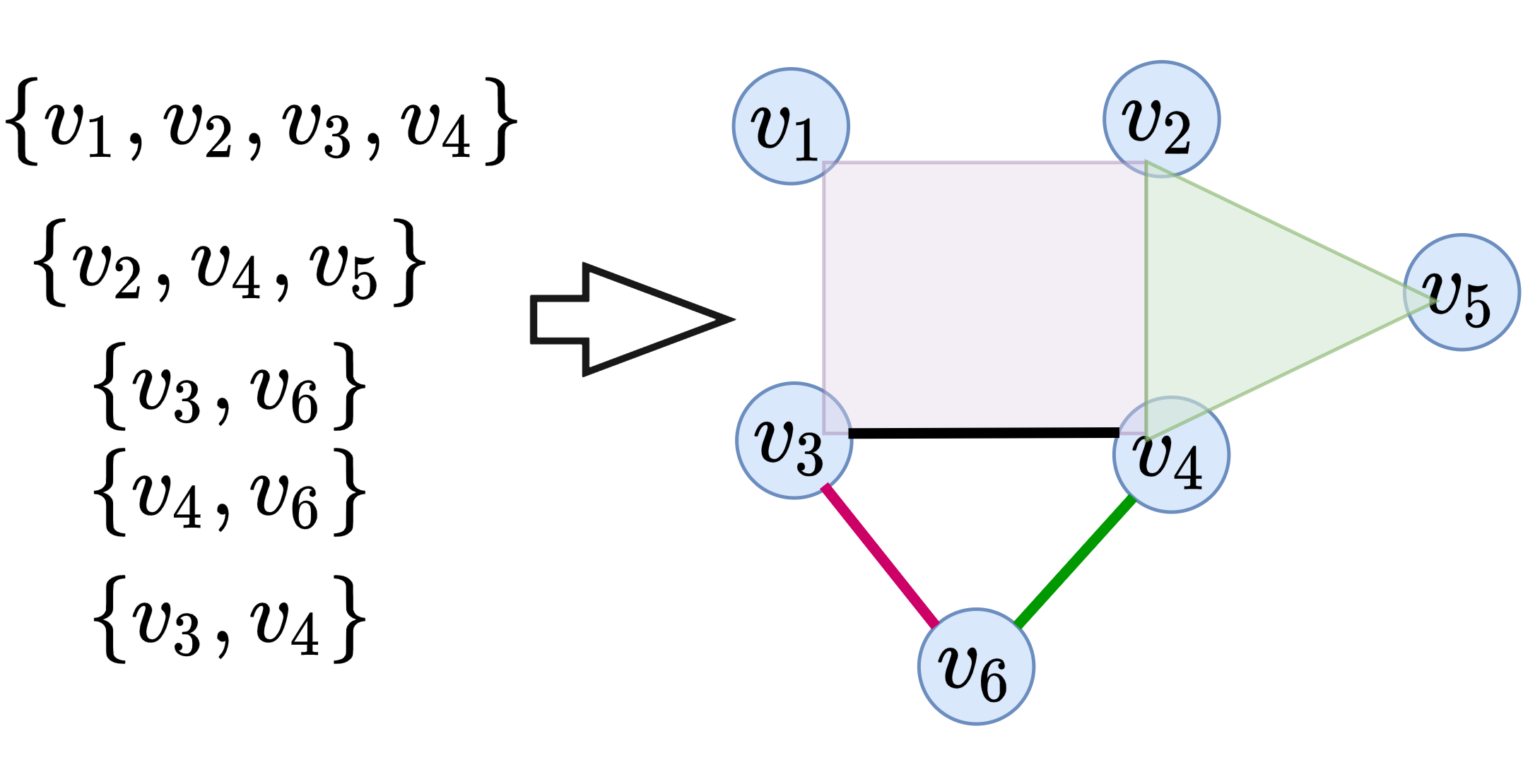}\label{fig:hypergraph1}}%
	\subfloat[][Hyperedges at time $t$]{\includegraphics[width=0.37\textwidth]{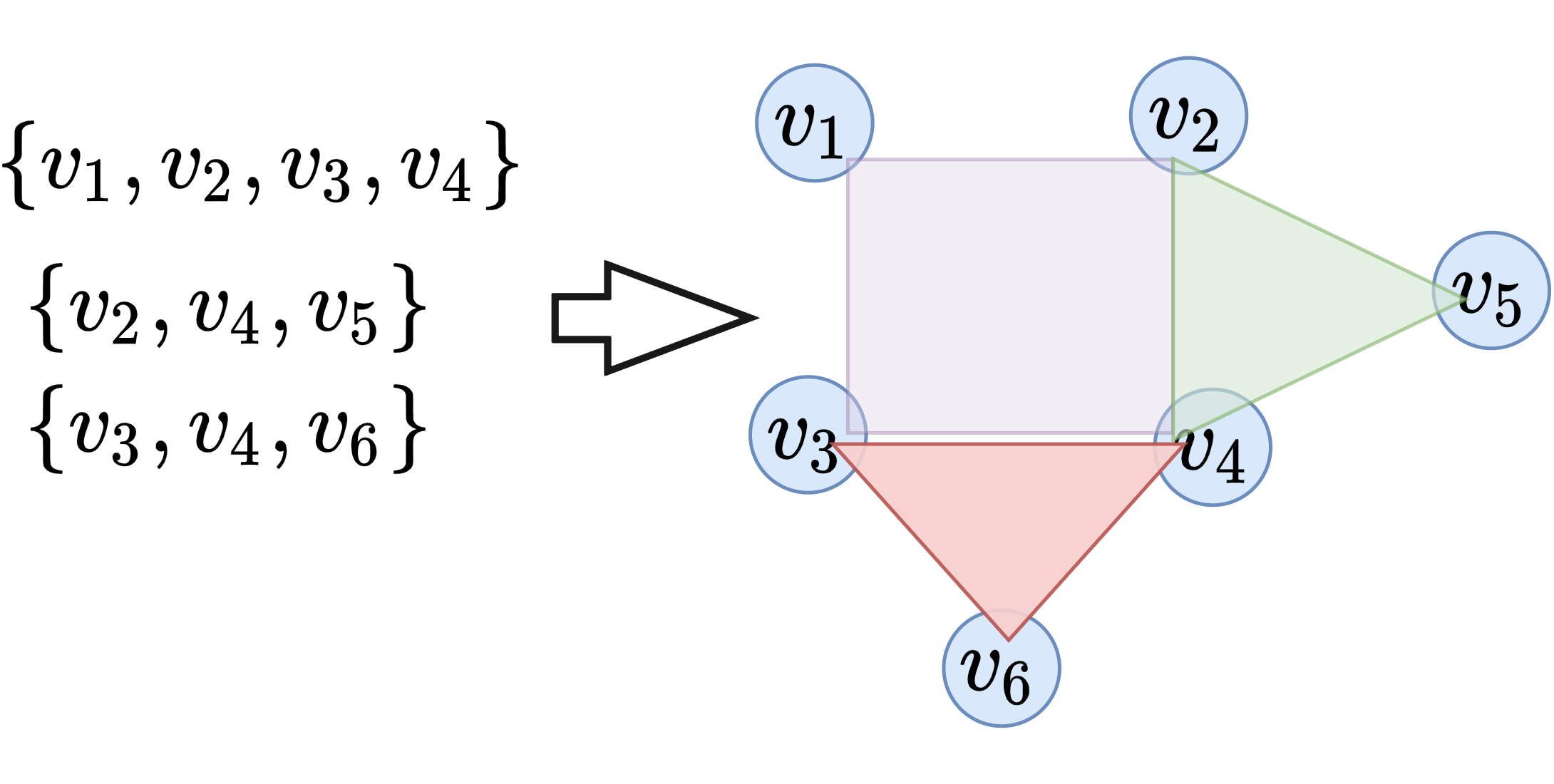}\label{fig:hypergraph2}}%
	\subfloat[][Approximate pairwise graph for Hypergraphs in \ref{fig:hypergraph1} and \ref{fig:hypergraph2}]{\includegraphics[width=0.2\textwidth]{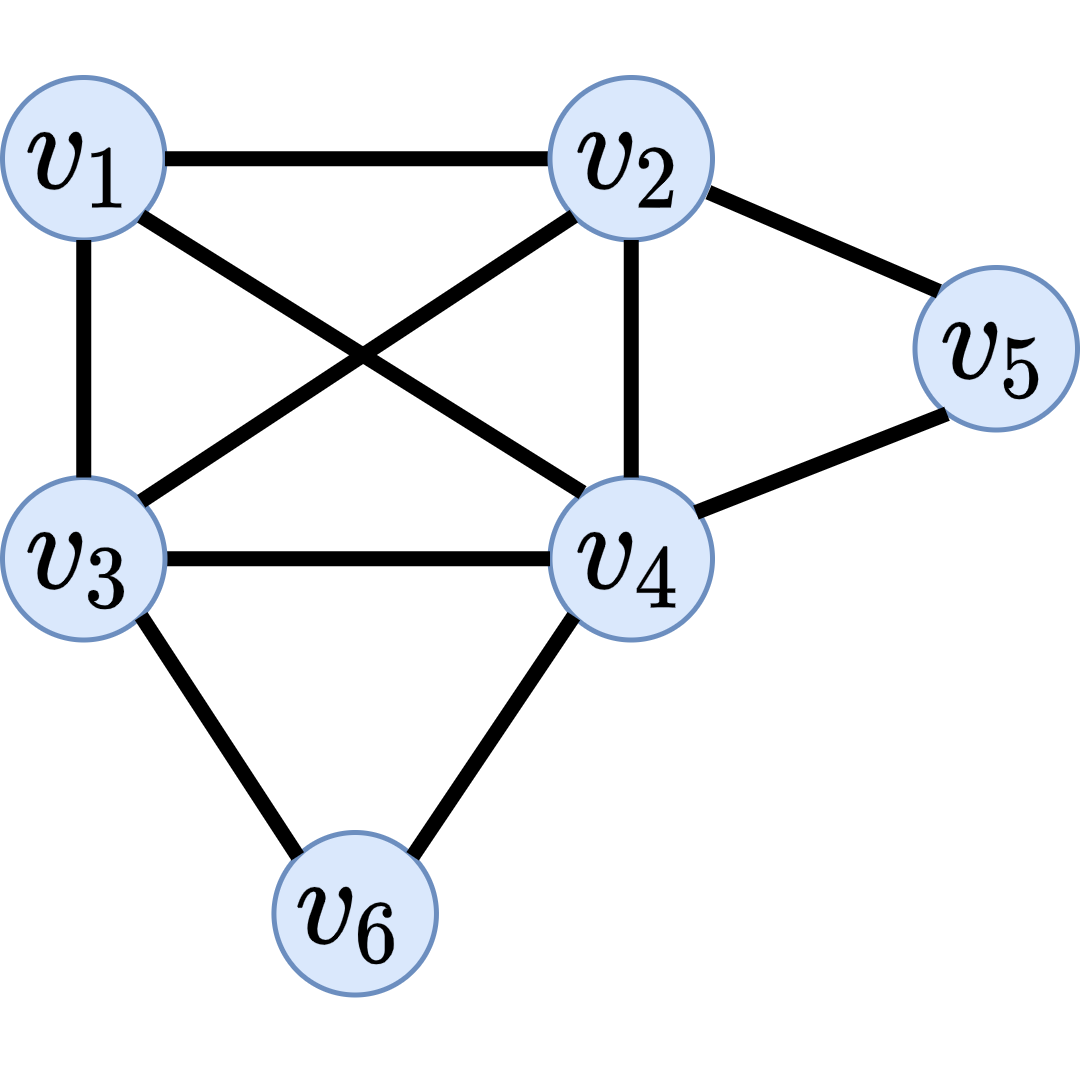}\label{fig:projectedgraph}}
	\caption{Higher-order interactions at time $t$ are shown as hyperedges in Figures~\ref{fig:hypergraph1} and~\ref{fig:hypergraph2}.  Here, hyperedges are represented by geometric shapes with their ends/corners showing the nodes and color showing their identity. One can see two different hypergraphs having the same projected graph in Figure~\ref{fig:projectedgraph}. This demonstrate a need for a technique that predicts hyperedges without approximating the hypergraph with a pairwise graph.}
	\label{fig:decomposing hyperedge to pairwise}
\end{figure*}

However, most real-world interactions are more complex than just pairwise interactions. For example, a person can have multiple items in a single shopping order, a group of people can co-author an article,  mutual funds have stocks of various companies, and so on. These types of problems have been studied using hypergraphs, and there have been many theoretical advances have taken place \citep{2017:GhoshdastidarDukkipati:2017:ConsistencyOfSpectralHypergraphPartitioningUnderPlantedPartitionModel,2017:GhoshdastidarDukkipati:2017:UniformHypergraphPartitioning}.

A common technique employed to deal with this problem is to approximate these multiway interactions with pairwise interactions, which amounts to approximating hypergraphs with graphs. This leads to enormous information loss, as demonstrated in Figure~\ref{fig:decomposing hyperedge to pairwise}. In this example, two different kinds of interaction between nodes $\{v_1, v_2, v_3, v_4, v_5, v_5\}$ have the same pairwise interaction graph. Further, it is impossible to infer the original interactions once they are projected into a pairwise graph. 

In this paper, we address the problem of forecasting higher-order interactions as hyperedge events using TPP. We define a conditional intensity function on each hyperedge that takes node representations as inputs. Since each hyperedge can have a variable number of nodes, we use a self attention-based architecture for hyperedge encoding. Since nodes evolve as they interact, it is essential to employ dynamic node representations. In earlier works on dynamic networks~\citep{trivediEtAL:2019:DyRep:LearningRepresentationsoverDynamicGraphs, Dai:EtAL:2016:Deep_coevolutionary_network:_Embedding_user_and_item_features_for_recommendation, Cao:EtAL:2021:Deep_Structural_Point_Process_for_Learning_Temporal_Interaction_Networks}, node embeddings are updated based on the node embedding of the other node in the interaction. For example, consider edge event $(v_a, v_b)$ occurring at time $t$. To update node embeddings of $v_a$, we use the node embeddings of $v_b$ and vice versa. However, hyperedge events have a variable number of nodes (Figure~\ref{fig:decomposing hyperedge to pairwise}), so the techniques developed for pairwise interaction are not directly applicable to higher-order interaction. Hence, to address this, we use a self-attention-based encoding for node updates with parameters shared with the hyperlink prediction model.

\paragraph{Contributions.} We propose a model called \textit{Hypergraph Dynamic Hyperedge} (HGDHE) to model higher-order interactions as hyperedge events in a dynamic Hypergraph. Further, one can notice that these interactions will not always be between a homogeneous set of nodes. Hence, we also developed a bipartite hyperedge variant of our model called \textit{Hypergraph Bipartite Dynamic Hyperedge} (HGBDHE). This will help in modeling interactions between two different types of nodes. Our contributions are as follows. 
\textbf{(1)} A temporal point process framework for hyperedge modeling that can forecast the type and time of interaction;
\textbf{(2)} A model for representation learning for higher-order interaction data;
\textbf{(3)} Extensive experiments on real-world datasets on both homogeneous and bipartite hyperedges;
\textbf{(4)} Empirical results on performance gain obtained when we use hyperedges instead of pairwise modeling; and
\textbf{(5)} Empirical results on performance gain are obtained when we use dynamic models instead of static models.
\paragraph{Related Works.} Earlier works in modeling temporal information into networks can be categorized as (i) discrete-time models and (ii) continuous-time models. In discrete-time models, time is discretized into bins of equal size, and recurrent neural network-based models are used for modeling temporal evolution~\citep{Gupta:EtAL:2019:A_Generative_Model_for_Dynamic_Networks_with_Applications,Gracious:EtAL:2021:Neural_Latent_Space_Model_for_Dynamic_Networks_and_Temporal_Knowledge_Graphs}. Since discretization results in information loss and selecting bin size is a difficult task, the recent focus has been on continuous-time models~\citep{Kumar:EtAL:2019:Predicting_Dynamic_Embedding_Trajectory_in_Temporal_Interaction_Networks,Xu:EtAL:2020:Inductive_representation_learning_on_temporal_graphs, Rossi:EtAL:2020:Temporal_graph_networks_for_deep_learning_on_dynamic_graphs}.
Unlike these discrete-time models, TPP-based continuous-time models can predict both dynamic interaction and time of interaction. Recently neural network-based TPP has been proposed to model dynamic interaction. However, these works approximate higher-order interactions with pairwise interactions~\citep{Dai:EtAL:2016:Deep_coevolutionary_network:_Embedding_user_and_item_features_for_recommendation,trivediEtAL:2019:DyRep:LearningRepresentationsoverDynamicGraphs,Cao:EtAL:2021:Deep_Structural_Point_Process_for_Learning_Temporal_Interaction_Networks} 
It has been shown in Hyper-SAGNN~\citep{Zhang:EtAL:2019:Hyper-SAGNN} that directly modeling the higher-order interaction will result in better performance than decomposing them into pairwise interactions. 


Higher-order interaction between nodes can be modeled as link prediction in a hypergraph. Earlier works use matrix completion techniques to predict hyperedge. Coordinated Matrix Minimization (CMM)~\citep{Zhang:EtAL:2018:Beyond_Link_Prediction:_Predicting_Hyperlinks_in_Adjacency_Space}, infer the missing hyperedges in the network by modeling adjacency matrix using a non-negative matrix factorization based latent representation. Recent works mainly concentrate on neural network-based scoring functions as they perform better than matrix completion-based techniques and are easier to train. Hyperpath~\citep{Huang:EtAL:2019:Hyper-Path-Based_Representation_Learning_for_Hyper-Networks} model's hyperedge as a tuple and uses a neural network-based scoring function to predict links. This method cannot model higher-order interactions as it expects the hyperedge size to be uniform for all edges. 
HyperSANN~\citep{Zhang:EtAL:2019:Hyper-SAGNN} uses a self-attention-based architecture for predicting hyperlinks. It can learn node embeddings and predict non-uniform hyperlinks. However, it is a static model and cannot model the dynamic nature of hyperedges. The work presented in this paper is the first work that uses TPPs to forecast hyperedges when networks are evolving with time. 


\section{Dynamic Hyperedge Forecasting}
\label{sec:dynamic_hyperedge_forecasting}
Given a set of nodes $\mathcal{V} = \{v_1, v_2, \ldots, v_{|\mathcal{V}|}\}$, an interaction event between a subset of these nodes is modeled as a hyperedge ($h_i$) with the time of interaction as its edge attribute. Here, $h_i \in \mathcal{H} $ is a subset of nodes in $\mathcal{V}$ and $\mathcal{H}$ is set of all valid combination of nodes. Given the historical events  $\mathcal{E}(t_m)= \{ (h_1, t_1), , \ldots, (h_m, t_m)\}$ till time $t_m$, aim is to forecast future hyperedge $h_i$ at time $t > t_m$. 

\subsection{Hyperedge Event Modeling} \label{sec:hyperedgeventmodelling}
Given a hyperedge $h = \{v_1, v_2, \ldots, v_k\}$, the probability of $h$ occurring at time $t$ can be modeled using TPP with conditional intensity $\lambda_h (t)$ as 
\begin{align} \label{eq:probabilityofevent}
	p_h (t)  =  \lambda_h (t) \mathcal{S}_h(t),
\end{align}
where $\mathcal{S}_h(t)$ is survival function that denotes probability that no event happened during the   interval $[t_{h}^p,t)$ for hyp eredge $h$. This is defined as 
\begin{align} \label{eq:survivalfunction}
	\mathcal{S}_h(t) = \exp{ \left( \int_{t_{h}^p}^t-\lambda_h (\tau) \ud \tau  \right) }, 
\end{align}  
where ${t_{h}^p} = \Max_{v \in h}  t_{v}^p$ and $t_{v}^p$ denotes the most recent interaction time of node $v$ in $h$. We provide a more detailed background explanation of TPP in Appendix~\ref{sec:Temporal_Point_Process}. The conditional intensity function $\lambda_h (t) $ is parameterized by defining a positive function over the embeddings of nodes in $h$ as, 
\begin{align} \label{eq:intensity}
	\lambda_h (t) = f( \bm{v_1} (t) , \bm{v_2} (t), \ldots, \bm{v_k} (t)  )    .
\end{align}
Here $f( . ) \geq 0 $ can be realized by neural network for hyperedge events, and $\bm{v_i}(t) \in \mathbb{R}^d$ is the node embeddings at time $t$ for node $v_i$. We follow the same architecture of the hyperedge modeling technique as Hyper-SAGNN~\citep{Zhang:EtAL:2019:Hyper-SAGNN} with a final softplus layer as explained in Appendix~\ref{sec:lambda_homogeneous}. In this paper, we have also developed baseline models using piece-wise constant node embeddings with intensity defined by Rayleigh Process to create an equivalent model for DeepCoevolve~\citep{Dai:EtAL:2016:Deep_coevolutionary_network:_Embedding_user_and_item_features_for_recommendation} in hyperedge interactions, as shown below, 
\begin{align} \label{eq:rayleighintensity}
	\lambda_h (t) =  f\Big( \bm{v_1} (t_h^p) , \bm{v_2} (t_{h}^p), \ldots, \bm{v_k} (t_{h}^p) \Big) (t - t_h^p ) .
\end{align}
This formulation will give us a closed-form solution for survival function, thereby for the probability of the event. Otherwise, integration has to be approximated by sampling.

\subsection{Dynamic Node Representation} 
\label{sec:DynamicNodeRepresentation} 
For each node in the network, we learn a low dimensional embedding $\bm{v}(t) \in \mathbb{R}^d$ that changes with time. It is done through three stages,  i) Temporal drift, ii) History aggregation, and iii) Interaction update, as 

\begin{align} \label{eq:dynamic_node_embeddings_hgcn}
	\bm{v}(t)= \tanh\Big( \underbrace{\bm{W_{0}} \bm{v} ( {t_v^p}^{+} )}_{\textbf{Interaction Update}} & + \underbrace{\bm{W_{1}} \bm{\Phi}( t - t_v^p )}_{\textbf{Temporal Drift}} + \nonumber \\
	& \underbrace{\bm{W_{2}} \bm{v^{s}} (t_{i-1})}_{\textbf{History Aggregation}}  + \bm{b_{0} }  \Big).
\end{align}
Here, $\bm{W_0}, \bm{W_1}, \bm{W_2} \in \mathbb{R}^{d \times d}, \bm{b_0} \in \mathbb{R}^d$ are learnable parameters, $\bm{v}( {t_v^p}^{+} )$ is the node embedding just after previous interaction for node $v$ at time $t_v^p$ and $t_{i-1}$ is the last event time, $t_{i-1} < t$.

\paragraph{\textit{Temporal drift.}} This term models the inter-event evolution of a node with time. For a node $v$ with previous event time $t^p_v$, the drift in embedding at time $t$ is modelled by $\bm{W_{1}} \bm{\Phi}( t - t_v^p )$, where $\bm{\Phi}(t) \in \mathbb{R}^d$ is Fourier time features~\citep{Xu:EtAL:2020:Inductive_representation_learning_on_temporal_graphs,Cao:EtAL:2021:Deep_Structural_Point_Process_for_Learning_Temporal_Interaction_Networks} defined as  
$
\bm{\Phi}(t) = [\cos(\omega_1 t + \theta_1) , \ldots, \mathrm{cos}(\omega_d t + \theta_d)  ].     
$
Here, $\{\omega_i \}_{i=1}^d$, and  $\{\theta_i \}_{i=1}^d$ are learnable parameters.


\paragraph{\textit{History aggregation.}} This stage uses hypergraph convolution-based feature aggregation to incorporate the effect of past events. For this, we will construct a hypergraph using the past $M$ events, $\{ (h_{i-M}, t_{i-M} ), \ldots, (h_{i-1}, t_{i-1} )\}$, and uses its incidence matrix $\mathbf{H}(t_{i-1}) \in \mathbb{R}^{|V| \times M }$ to apply hypergraph convolution as 
$
[\bm{v_1^s} (t_{i-1}), \ldots,  \bm{v_{|\mathcal{V}|}^s}(t_{i-1})]= \mathrm{HGNN} \left( \mathbf{H}(t_{i-1}), \left[\bm{v_1} ( {t_{v_1}^p}^{+} ), \ldots, \bm{v_1} ( {t_{v_{|\mathcal{V}|}}^p}^{+} ) \right] \right).
$
Here, $\mathrm{HGNN}$ is a hypergraph graph convolution defined as in~\citep{Bai:EtAL:2021:Hypergraph_convolution_and_hypergraph_attention}.


\paragraph{\textit{Interaction update.}} When a node $v$ is involved in an interaction $h$, it is influenced by the nodes it interacts within $h$. For extracting features of interaction, we use the dynamic embedding $\bm{d_v^h} $ calculated as a function of embeddings of nodes $h-\{v\}$ at time t. The architecture of calculating this is shared with conditional intensity function as shown in Equation~\ref{eq:dynamic_embeddings_homogeneous} in Appendix~\ref{sec:lambda_homogeneous}.The entire update equation is $
\bm{v}({t}^{+}) =  \mathrm{tanh}(\bm{W_{3}}\bm{v}( {t_v^p}^{+} ) + \bm{W_{4}}\bm{\Phi}( t - t_v^p ) + \bm{W_5} \bm{d_v^h}  + \bm{b_{1}} ).
$
Here, $\bm{W_3}, \bm{W_4}, \bm{W_5} \in \mathbb{R}^{d \times d} $ and $\bm{b_1} \in \mathbb{R}^d$ are learnable parameters. If node $v$ is involved in multiple hyperedge events $\{h_{i}, h_{i+1}, \ldots h_{i+L} \}$, then we will take the mean of dynamic embeddings from all of the hyperedges. Here, $L$ is the number of concurrent hyperedges.  


\section{Learning Procedure} 
\label{sec:Learning_Procedure}
\subsection{Loss Function} \label{sec:Loss_Function}
Once intensity parameterization is fixed for temporal point process as in Equation~\ref{eq:intensity} or~\ref{eq:rayleighintensity}, the likelihood for hyperedge events $\mathcal{E}(T)= \{ (h_1, t_1),  \ldots, (h_m, t_m)\}$ occurring in an interval $[0, T]$ can be modeled as, 
$
p(\mathcal{E}(T)) = \prod_{i=1}^{m}  p_{h_i}(t_i) \prod_{h \in \mathcal{H}} \mathcal{S}_h(t^\ell_h, T) .
$   
Here, $p_{h_i}(t_i)$ is the probability of hyperedge event $h_i$ occurring at time $t_i$ as defined in Equation~\ref{eq:probabilityofevent}, $\mathcal{S}_h(t^\ell_h, T)$ is the probability that no event occurred for hyperedge $h$ for the interval $[t^\ell_h, T]$, and $t^\ell_h$ is the last time occurrence for $h$ ($t^\ell_h=0$, when no event of $h$ is observed). The loss for learning parameters can be found by taking the negative of log-likelihood as, 
$
\mathcal{L} = - \sum_{i=1}^{m}\log ( \lambda_{h_i}(t_i)) + \sum_{ h \in \mathcal{H}} \int_{0}^T \lambda_{h} (t) \ud t . 
$    
Here, the first term corresponds to the sum of the negative log intensity of occurred events. The second term corresponds to the sum of intensities of all events. The following happens when we minimize the loss, the intensity rate of occurring events increases due to minimization of the first term, and intensity rates of events not occur decrease due to minimization of the second term. However, directly implementing this equation is computationally inefficient as $|\mathcal{H}| \leq 2^{|\mathcal{V}|}$ is very large. Further, the integration in the second term does not always have a closed-form expression. In the next section, we will give this model a computationally efficient mini-batch training procedure. 

\subsection{Mini-Batch Loss} 
\label{sec:Mini-Batch_Loss}
We divide the event sequences into independent segments to make backpropagation through time feasible, as done in previous works on pair-wise dynamic networks~\citep{Dai:EtAL:2016:Deep_coevolutionary_network:_Embedding_user_and_item_features_for_recommendation}. Then the loss for each segment is calculated as follows, for each $(h_i, t_i$) in the segment $\mathcal{E}_M =\{ (h_1, t_1), \ldots, (h_M, t_M)\}$, we use Monte-Carlo integration to find the $\log$ survival term of $p_{h_i}(t_i)$. Then the negative log-likelihood for that event is,
\begin{align} \label{eq:samplenegativeloglikelihood}
	\mathcal{T}^s &= \{t^s_j \}_{j=1}^N \leftarrow \mathrm{Uniform}(t_{i-1}, t_i, N ) \nonumber \\ 
	\mathcal{L}_{h_i} &= - \log(\lambda_{h_i}(t_i)) + \sum_{j=2}^N (t^s_j - t^s_{j-1}) \lambda_{h_i}(t^s_j). 
\end{align}
Here, $\mathcal{T}^s $ is the set of uniformly sampled time points from the interval $[t_{i-1}, t_i]$. Then to consider the interactions events $h \in \mathcal{H}$ that were not observed during the above period, we sample some negative hyperedges for each interaction event $(h_i, t_i)$ as described below,
\begin{enumerate}
	\item Choose the size of negative hyperedge $k$ based on a categorical distribution over hyperedge sizes observed in the training data. Here, parameters of the categorical distribution are learned from the training dataset.
	\item Sample $\min( \left\lceil k/2\right\rceil , |h_i| )$ nodes from the hyperedge $h_i$ and rest of the nodes from $\mathcal{V}- h_i$. This strategy will avoid trivial negative samples.
\end{enumerate}
Following the above steps, we sample $\mathcal{H}^{n}_i = \{  h^n_1, \ldots, h^n_{\mathcal{B}}\}$ negative hyperedges, and for each of them, we calculate the negative log-likelihood for events not happening using Monte-Carlo integration. Then Equation~\ref{eq:samplenegativeloglikelihood} becomes, 
$$
\mathcal{L}_{h_i} = - \log(\lambda_{h_i}(t_i)) + \sum_{h \in \mathcal{H}^{n}_i \cup \{h_i\} }\sum_{j=2}^N (t^s_j - t^s_{j-1}) \lambda_{h}(t^s_j). 
$$ 
The final mini-batch loss is calculated by summing all $\mathcal{L}_{h_i}$ for the events $(h_i, t_i)$ in $\mathcal{E}_M$, $\sum_{i=1}^{M}\mathcal{L}_{h_i}$. Then the gradients are backpropagated for this loss, and the above training procedure is repeated in the next segment.

\section{Extending to Bipartite Hyperedges} 
\label{sec:bipartite_hyperedge_extension}
A higher-order interaction between nodes of two different type can be represented as a bipartite hyperedge $h=( \{v_1, \ldots , v_k\},  \{v_{1^{'}}, \ldots , v_{k^{'}}\} )$. Here,  $\{v_1, \ldots , v_k\} \in \mathcal{H}$ is the left hyperedge with nodes from set $\mathcal{V}$ , $\{v_{1^{'}}, \ldots , v_{k^{'}}\} \in \mathcal{H}^{'}$ is the right hyperedge with nodes from set $\mathcal{V}^{'}$, and $\mathcal{V} \cap \mathcal{V^{'}}=\emptyset$. More details on these types of higher-order heterogeneous graphs can be found in the recent work~\citep{Sharma:2021:CATSET}. For defining conditional intensity function, similar to homogenous hyperedges, we define $\lambda_h (t) =  f( \{v_1, \ldots , v_k\},  \{v_{1^{'}}, \ldots , v_{k^{'}}\} )$. Here $f( . ) \geq 0 $ is defined by a neural network, and $\bm{v_i}(t), \bm{v_{i^{'}}} (t)$ are node embeddings of the nodes in $h$. We follow the same architecture of CATSETMAT~\citep{Sharma:2021:CATSET} for defining $f( . )$ as explained in Appendix~\ref{sec:lambda_bipartite}. 

Now to learn good dynamic representation for nodes, we follow the same architecture explained in Section~\ref{sec:DynamicNodeRepresentation}.  But, the left and right hyperedges have their own set of parameters for the \texttt{Temporal drift}, \texttt{History aggregation}, and \texttt{Interaction update} stages. Further, for dynamic embeddings in the \texttt{Interaction update} stage, nodes in the left hyperedge have it as a function of embeddings of nodes in the right hyperedge, and vice versa. This function shares its parameters with the conditional intensity function, as shown in Equation~\ref{eq:bipartite_dynamic_embeddings} in  Appendix~\ref{sec:lambda_bipartite}.

For learning parameters, we follow the same procedure as that of homogenous hyperedges as explained in Section~\ref{sec:Learning_Procedure}, except the negative sampling is done differently. We keep the left or right hyperedge fixed for generating negative samples and add a corrupted hyperedge on the other side. For example, for generating left hyperedge negative samples, we replace the right hyperedge by selecting a random subset of nodes from the right node-set $\mathcal{V}^{'}$. The size of the corrupted hyperedge is selected based on the categorical distribution of sizes of the right hyperedge in the training set.


\begin{table}
	\small 
	\setlength{\tabcolsep}{5pt}
	\begin{tabular}{lrrrrr}
		\toprule
		\textbf{Datasets}  &  $|\mathcal{V}|$ & $|\mathcal{V}^{'}|$ & $|\mathcal{E}(T)|$ & $|\mathcal{H}|$ & $|\mathcal{H}^{'}|$ \\
		\midrule 
		\textbf{email-Enron} & 143 &  N/A  & 10,883 & 1,542 & N/A \\
		\textbf{email-Eu} & 998 & N/A & 234,760 & 25,791 & N/A \\
		\textbf{congress-bills}  & 1,718 & N/A & 260,851 & 85,082 & N/A \\
		\textbf{NDC-classes}   & 1,161 & N/A & 49,724 & 1,222 & N/A \\
		\textbf{NDC-sub} & 5,311 & N/A & 112,405 & 10,025 & N/A  \\ 
		\textbf{CastGenre} & 5,763 & 20 & 12,295 & 11,665 & 1,078 \\
		\textbf{CastKeyword} & 3,998 & 1,953 & 13,826 & 12,365 & 10,737 \\
		\textbf{CastCrew} &  5,763 & 4,541 & 12,295 & 11,665 & 9,451 \\ 
		\bottomrule
	\end{tabular}
	\caption{Datasets used for Homogeneous and Bipartite Hyperedges along with their vital statistics.  }
	\label{tab:datasets}
\end{table}

\section{Experimental Settings}
\subsection{Datasets}
All the datasets for homogeneous hyperedge interactions  in Table~\ref{tab:datasets} are taken from work~\citep{Benson:EtAL:2018:Simplicial_closure_and_higher-order_link_prediction}\footnote{https://www.cs.cornell.edu/~arb/data/}. The bipartite hyperedges interactions are prepared from Kaggle's 
The Movies Dataset~\footnote{https://www.kaggle.com/datasets/rounakbanik/the-movies-dataset}. A detailed explanation of all the datasets is given in Appendix~\ref{sec:datasets}. For homogeneous hypergraphs,  $|\mathcal{V}^{'}|$ is not applicable (N/A) as there is only one type of node, and $|\mathcal{H}^{'}|$ is N/A as it is not directed.


\begin{table}
	\small
	\setlength{\tabcolsep}{5pt}
	\begin{center}
		\begin{tabular}{l | c | c | c| c | c}
			\toprule
			\textbf{Methods} & \textbf{T.D.} & \textbf{H.G.} & \textbf{I.U.} & \textbf{Hyperedge} & \textbf{Bipartite} \\ 
			\midrule
			RHE & \xmark & \xmark & \xmark & \cmark  & \xmark \\
			RDHE  & \cmark & \xmark & \cmark & \cmark & \xmark \\
			\midrule
			DE-drift   & \cmark & \xmark & \xmark & \xmark & \xmark \\
			DE  &  \cmark & \xmark & \cmark & \xmark & \xmark \\
			\midrule
			DHE-drift   & \cmark & \xmark & \xmark & \cmark & \xmark \\
			DHE & \cmark & \xmark & \cmark & \cmark & \xmark \\
			\midrule
			HGDHE-hist & \cmark & \cmark & \xmark & \cmark & \xmark \\
			HGDHE  & \cmark & \cmark & \cmark & \cmark & \xmark \\
			\midrule
			BDE  & \cmark & \xmark & \cmark & \xmark & \cmark \\
			\midrule
			BDHE  & \cmark & \xmark & \cmark & \cmark & \cmark \\
			\midrule
			HGBDHE  & \cmark & \cmark & \cmark & \cmark & \cmark \\
			\bottomrule
		\end{tabular}
		\caption{Models and their properties. Here, \cmark\ indicates the usage of that property, and \xmark\ indicates the absence of that property. Here, \texttt{T.D.} corresponds to \texttt{Temporal drift}, \texttt{H.G} corresponds to \texttt{History aggregation}, and \texttt{I.U.} corresponds to \texttt{Interaction update}.}
		\label{tab:model_propertities}
	\end{center}
\end{table}



\subsection{Baselines}
In Table~\ref{tab:model_propertities}, we have compared the properties of the baseline models we created against the proposed models HGDHE for the homogenous hyperedge interactions and HGBDHE for the bipartite hyperedge interactions. Here, models \textit{Dynamic Edge} (DE) and \textit{Dynamic Edge-drift} (DE-drift) use pairwise edge models instead of hyperedge interaction. Similarly, model \textit{Bipartite Dynamic Edge} (BDE) uses bipartite pairwise edges to model bipartite hyperedge interaction. Further, in models \textit{Rayleigh Hyperedge} (RHE) and \textit{Rayleigh Dynamic Hyperedge} (RDHE), conditional intensity is modeled as the Rayleigh process as in Equation~\ref{eq:rayleighintensity}, and duration predictions are made using the closed-form expression in Equation~\ref{eq:durationpredictionrayliegh}. \textit{Dynamic Hyperedge} (DHE) is the model that uses hyperedges for predicting higher-order interactions and has the same dynamic node presentation as of DE model. So, in our studies, we will compare these two models to claim and establish the advantage of hyperedge modeling over pairwise modeling.  Similarly, \textit{Dynamic Hyperedge-drift} (DHE-drift)  is the hyperedge model version of the DE-drift model. In the case of bipartite hyperedge interactions, we will compare models \textit{Bipartite Dynamic Hyperedge} (BDHE) and BDE. A more detailed description of baselines can be found in Appendix~\ref{appendix:baselines}.

Apart from the models mentioned above, we also provide comparisons against the current state-of-the-art approaches for temporal pairwise graphs DyRep \cite{trivediEtAL:2019:DyRep:LearningRepresentationsoverDynamicGraphs}, TGAT \cite{Xu:EtAL:2020:Inductive_representation_learning_on_temporal_graphs}, and TGN \cite{Rossi:EtAL:2020:Temporal_graph_networks_for_deep_learning_on_dynamic_graphs}.  In this, DyRep uses the temporal point process framework for model edges and can predict both type and time of interaction. The other two models, TGAT and TGN, use a continuous time dynamic graph neural network for modeling edges and can only predict the type of interaction occurring at a particular time.

\subsection{Prediction Tasks}
Using temporal point process models, one can predict both the next event type and the time of the event. The following are the equivalent tasks in our settings.

\paragraph{\textit{Interaction type prediction.}} The type of interaction that occurs at time $t$ can be predicted by finding the $h_i$ with the maximum intensity value at that time, as shown below,
\begin{align} \label{eq:interactionprediction}
	\hat{h} = \argmax_{h_i} \lambda_{h_i}(t) .
\end{align}
\paragraph{\textit{Interaction duration prediction.}} For interaction $h$ occurred at time $t^p_h$, to predict the duration for future interaction, we have to calculate the expected time $t$ with respect the conditional distribution $p_h(t)$ in Equation \ref{eq:probabilityofevent}, 
\begin{align} \label{eq:durationprediction}
	\hat{t} = \int_{t^p_h}^{\infty} (t - t^p_h ) p_h(t) \ud t . 
\end{align}

If $\lambda(t)$ is modeled using a Rayleigh process as in Equation \ref{eq:rayleighintensity}, we calculate the $\hat{t}$ in close form as, 
\begin{align} \label{eq:durationpredictionrayliegh}
	\hat{t} = \sqrt{\frac{\pi }{2 \exp{f( \bm{v_1} (t_h^p) , \bm{v_2} (t_{h}^p), \ldots, \bm{v_k} (t_{h}^p) )}}}.
\end{align}
Otherwise, we have to compute the integration by sampling.


\subsection{Metrics of Evaluation}
\paragraph{\textit{Mean Reciprocal Rank (MRR).}} We use this for evaluating the performance of interaction prediction at time $t$. For finding this, we find the reciprocal of rank ($r_i$) of the true hyperedge against candidate negative hyperedge in descending order of $\lambda_h (t)$ and then average them for all samples in the test set, $MRR = \frac{1}{N}\sum_{i=1}^{N} \frac{1}{r_i + 1}$. Here, better-performing models have higher MRR values.

\paragraph{\textit{Mean Absolute Error (MAE) .}} We use this for evaluating the performance of interaction duration prediction, $MAE = \frac{1}{N}\sum_{i=1}^N |\hat{t}_i - t^{true}_i |$. Here, better-performing models have lower MAE values. 
\subsection{Parameter Settings}
\label{sec:parameter_settings}
For all experiments, we use the learning rate of $0.001$, the embedding size $d$ is fixed at $64$ for homogeneous hyperedges and bipartite hyperedges, the batch size $M$ is fixed as 128, and the negative sampling is fixed as $\mathcal{B}=20$, and the training is done for $100$ epochs. For the Monte Carlo estimate of $\log$ of survival probability in Section~\ref{sec:Mini-Batch_Loss}, we use $N=20$ for datasets email-Enron, email-EU, and NDC-classes, $N=5$ for NDC-sub and congress-bills datasets. The choice of $N$ is made by considering memory constraints. All models are implemented PyTorch~\citep{Adam:EtAL:2019:PyTorch}, and all training is done using its Adam~\citep{Kingma:EtAL:2015:Adam:_A_Method_for_Stochastic_Optimization} optimizer. For all datasets, we use the first $50\%$ of interactions for training, the next $25\%$ for validation, and the rest for testing. The details of the computational infrastructure used are provided in the Appendix~\ref{appendix:computational_infrastructure}.  All the reported scores are the average of ten randomized runs along with their standard deviation.

\begin{table*}
	\centering
	\resizebox{0.99\textwidth}{!}{
		\begin{tabular}{lcccccccccc }
			\toprule
			\multirow{2}{*}{\textbf{Methods}}  & \multicolumn{2}{c}{\textbf{email-Enron}} &  \multicolumn{2}{c}{\textbf{email-Eu}}  & \multicolumn{2}{c}{\textbf{congress-bills}}  &\multicolumn{2}{c}{\textbf{NDC-classes}}  & \multicolumn{2}{c}{\textbf{NDC-sub}}  \\
			\cmidrule{2-11}
			& \textbf{MRR} & \textbf{MAE} & \textbf{MRR} & \textbf{MAE} & \textbf{MRR} & \textbf{MAE} & \textbf{MRR} & \textbf{MAE} & \textbf{MRR} & \textbf{MAE}\\
			\midrule
			DyRep$^a$ & $42.08 \pm 1.14$ &  $23.74 \pm 1.68$ & $32.23 \pm 0.70 $ &  $22.77 \pm 0.87$ & $40.30 \pm 0.40$ & $6.83 \pm  0.12$  & $65.80 \pm 1.14$ & $3.50 \pm 0.20$ & $55.71 \pm 1.56$ & $5.35 \pm 0.43$ \\
			TGAT$^b$ & $52.60 \pm 0.95$ & $N/A$ & $57.53 \pm 0.89$ & $N/A$ &  $78.47 \pm 1.52$ & $N/A$ & $83.1 \pm 0.72$ & $N/A$ & $77.06 \pm 1.30$ & $N/A$ \\
			TGN$^c$ & $54.32 \pm 1.02 $& $N/A$ & $64.63 \pm 1.2$ & $N/A$ & $80.61 \pm 1.8$ & $N/A$ & $91.06 \pm0.81 $ & $N/A$   &  $79.33 \pm 1.90$ & $N/A$\\
			\midrule 
			RHE & $34.45 \pm 0.81$ & $127.19 \pm 12.17$  & $52.37 \pm 0.47$ & $26.43 \pm 0.47$ & $32.14 \pm 0.38$ & $54.13 \pm 9.60$ & $87.64 \pm 1.63$ & $8.98 \pm 1.09$ & $74.33 \pm 0.30$ & $4.66 \pm 0.15$ \\
			RDHE & $26.73 \pm 2.76$ & $34.22 \pm 0.49$ & $27.68 \pm 5.02$ & $17.54 \pm 0.68$ & $52.90 \pm 3.24$ & $2.44 \pm 0.22$ & $81.17 \pm 2.56$ & $6.29 \pm 0.91$ & $66.46 \pm 1.50$ &  $2.52 \pm 0.07$ \\
			\midrule
			DE-drift  & $30.84 \pm 0.29$ & $48.50 \pm 0.65$  & $43.47 \pm 1.69$ & $21.21 \pm 0.06$ & $43.27 \pm 0.23$ & $4.07 \pm 0.32$ & $60.64 \pm 0.19$  & $5.02 \pm 0.15$ & $64.90 \pm 0.29$ & $12.38 \pm 0.84$ \\
			DE &  $52.89 \pm 0.38 $ &  $16.52 \pm 3.14$  & $44.27 \pm 0.74$ & $19.37 \pm 1.27$ & $56.27 \pm 2.89$ & $2.65 \pm 0.42$ & $64.58 \pm 0.78$ & $3.60 \pm 0.38$ & $65.83 \pm 1.41$  & $13.82 \pm 1.93$ \\
			\midrule
			DHE-drift  & $50.25 \pm 1.65$ & $88.14 \pm 4.59$ & $58.38 \pm 0.20$ & $33.56 \pm 0.82$ & $84.50 \pm 0.13$ & $3.84 \pm 0.26$  & $88.68 \pm 0.71$ & $1.92 \pm 0.12$ & $79.31 \pm 0.41$ & $3.03 \pm 0.01$ \\
			DHE &  $60.57 \pm 1.97$  & $25.48 \pm 4.37$    & $64.03 \pm 2.22$ & $19.72 \pm 2.00$ & $\bm{92.21 \pm 0.19}$ & $1.87 \pm 0.22$ & $88.93 \pm 0.16$ & $1.84 \pm 0.21$ & $86.52 \pm 0.18$ & $3.49 \pm 0.14$\\
			\midrule
			HGDHE-hist & $\bm{65.26 \pm 1.24} $ & $18.25 \pm 0.43$ & $60.69 \pm 0.09$  & $24.66 \pm 0.77$ & $85.31 \pm 0.10$ & $3.44 \pm 0.34$  & $\bm{91.24 \pm 0.74}$ & $1.42 \pm 0.08$ & $80.73 \pm 0.15$ & $1.75 \pm 0.19$ \\
			HGDHE & $62.21 \pm 2.85$   & $\bm{16.12 \pm 1.45}$  &  $\bm{66.12 \pm 2.90}$  & $\bm{15.18 \pm 2.14}$ & $92.09 \pm 0.03$ & $\bm{1.65 \pm 0.06}$ & $91.01 \pm 0.35$ & $\bm{1.21 \pm 0.04}$ & $\bm{86.92 \pm 0.51}$  & $\bm{1.65 \pm 0.18 }$ \\
			\bottomrule
	\end{tabular}}
	\caption{Performance of dynamic homogeneous hyperedge forecasting in tasks of interaction type and interaction duration prediction. Here, interaction type prediction is evaluated using MRR in $\%$, and interaction duration prediction is evaluated using MAE. The proposed model HGDHE beats baseline models in almost all the settings. \textbf{Citations}: $^a$ \cite{trivediEtAL:2019:DyRep:LearningRepresentationsoverDynamicGraphs}, $^b$ \cite{Xu:EtAL:2020:Inductive_representation_learning_on_temporal_graphs}, $^c$  \cite{Rossi:EtAL:2020:Temporal_graph_networks_for_deep_learning_on_dynamic_graphs} }
	\label{tab:dynamiclinkprediction_1}
\end{table*}

\section{Results}
\subsection{Performance}
In Table~\ref{tab:dynamiclinkprediction_1}, one can see that the proposed model \textit{Hypergraph Dynamic Hyperedge} (HGDHE) performs better than baselines in almost all the settings. Further, it significantly outperforms RHE, which uses static, and RDHE, which uses piece-wise constant node embeddings. Even though these models have a closed form expression for the event probability and duration estimation $\hat{t}$, their performance is poor compared to HGDHE, which has conditional intensity as a function of its dynamic node representation. We can also see that by comparing baselines, DHE to RDHE and RHE to DHE-drift, the models DHE and DHE-drift perform better as they use dynamic node representations as input to the conditional intensity function. This is because those models have more expressiveness and do not assume a parametric form for $\lambda_h (t)$. 

Further, one can observe the advantage of using hyperedge for modeling higher-order interactions when comparing models DHE to DE and DHE-drift to DE-drift. This  comparison is important because those models use the same dynamic node presentation, but DHE uses hyperedge modeling, and DE uses pairwise edge modeling. The same applies to the comparison of DHE-drift to DE-drift. Between DHE and DE, there is an improvement in the MRR metric in interaction type prediction tasks for all datasets. There is a $38.4\%$ percentage gain in MRR for DHE compared to DE and a $40.3\%$ gain for DHE-drift compared to DE-drift. In the interaction duration prediction task, we can observe a significant reduction in MAE for all the datasets except for the email-Enron and email-Eu datasets. A similar observation can be made between DHE-drift and DE-drift models. This is because more than $70\%$ of interactions are pairwise in both of those datasets. Even though pairwise edges can achieve reasonable performance, we cannot identify the hyperedge among them if there are concurrent hyperedges with common nodes, as explained in Figure~\ref{fig:decomposing hyperedge to pairwise}. Hence, hyperedge models perform better than pairwise models for interaction type prediction.

\begin{table*}
	\centering
	\small
	\setlength{\tabcolsep}{2pt}
	\begin{tabular}{lcccccc}
		\toprule
		\multirow{2}{*}{\textbf{Methods}}  & \multicolumn{2}{c}{\textbf{CastGenre}} &  \multicolumn{2}{c}{\textbf{CastKeyword}} &  \multicolumn{2}{c}{\textbf{CastCrew}} \\
		\cmidrule{2-7}
		& \textbf{MRR} & \textbf{MAE} & \textbf{MRR} & \textbf{MAE} & \textbf{MRR} & \textbf{MAE}  \\
		\midrule
		BDE &   $23.55 \pm 0.75$ & $10.34 \pm 0.34$ & $13.61 \pm 0.14$  & $21.98 \pm 0.60$  & $13.61 \pm 0.24$ & $22.23 \pm 1.03$ \\
		\midrule
		DHE & $27.58 \pm 0.88$ & $\bm{2.88 \pm 0.43}$ & $36.18 \pm 1.34$ & $15.32 \pm 1.75$ & $26.03 \pm 1.75$ & $9.29 \pm 2.01$  \\
		BDHE &  $33.22 \pm 0.52$ & $2.91 \pm 0.26$  &  $38.77 \pm 1.69$ & $9.61 \pm 2.33$ & $37.29 \pm 2.65$ & $9.18 \pm 1.37$\\
		\midrule
		HGDHE & $27.59 \pm 1.60$ & $19.39 \pm 1.62 $ & $35.32 \pm 1.96$ & $22.64 \pm 1.50$ & $25.19 \pm 2.83$ & $8.85 \pm 2.07$  \\
		HGBDHE & $\bm{33.65 \pm 1.58}$ & $4.16 \pm 0.84$ & $\bm{41.32 \pm 1.74}$ & $\bm{9.27 \pm 1.67}$ & $\bm{42.77 \pm 2.00}$ & $\bm{8.77 \pm 1.68}$ \\
		\bottomrule
	\end{tabular}
	\caption{Performance of dynamic bipartite hyperedge forecasting in tasks of interaction type and interaction duration prediction. Proposed model HGBDHE beats baseline models in almost in all the settings.}
	\label{tab:dynamicdirectedlinkprediction}
\end{table*}
\paragraph{\textit{Comparison with previous works.}} In Table~\ref{tab:dynamiclinkprediction_1}, we have compared our model HGDHE against previous works on pairwise temporal graphs DyRep, TGAT, and TGN. Here, we can see that our model considerably outperforms previous works as they are not hyperedge prediction models. The margin of performance is higher for datasets congress-bills and NDC-Sub compared to email-Enron and email-Eu as they have more higher-order interactions than the latter. Further, the low performance of DyRep compared to other works TGAT and TGN is because of the staleness in embedding \cite{Kazemi:EtAL:2020:Representation_learning_for_dynamic_graphs:_A_survey} as it does not have a temporal drift stage to model evolve embedding during the interevent stage. The poor performance of TGAT and TGN is also because they use edge and node attributes along with graph neural networks to find the dynamic embeddings of the node. This information is not available for the datasets used in these experiments, so we initialized the node embeddings with random features and did not use edge features. 

\paragraph{\textit{Performance on Bipartite interactions.}} In Table~\ref{tab:dynamicdirectedlinkprediction}, one can observe that the proposed model \textit{Hypergraph Bipartite Dynamic Hyperedge} (HGBDHE) performs better than all the other baselines in almost all settings.  We can see that the models that use the bipartite property of the interaction perform better than their homogeneous counterparts. This can be inferred by the better performance of BDHE compared to DHE and HGBDHE compared to HGDHE. There is a gain of $23.6\%$ in MRR and a reduction of $12.4\%$ MAE for BDHE compared to DHE. Similarly, there is a gain of $36.2\%$ in MRR and a reduction of $46.1\%$ MAE for HGBDHE compared to HGDHE. This is because these models are more expressive and consider the bipartite nature of the interaction, as explained in Section~\ref{sec:bipartite_hyperedge_extension}. Similar to the case of homogeneous interactions, the pairwise model performs considerably poorer than the bipartite hyperedge models. It can be inferred from the poor performance of the BDE model, which uses a bipartite pairwise edge, compared to BDHE, which uses bipartite hyperedge for interaction modeling. Hence, one can conclude that bipartite hyperedge modeling represents data more accurately than pairwise modeling.

\subsection{Ablation Studies} 
\label{sec:ablation studies}

\paragraph{\textit{Effect of interaction update on performance.}} Here, we compare the performance of models that have this particular state in their update equation vs the models that do not. Firstly, we can observe a reduction of MAE by $24.7\%$ and a gain of $3.9\%$ in MRR for HGDHE compared to HGDHE-hist. Similarly, DHE outperforms DHE-drift considerably in the interaction type prediction task for all datasets. One can also observe a significant reduction in MAE for email-Enron and email-Eu datasets. Between RHE and RDHE, one can see that RDHE has a performance gain in interaction duration prediction in all datasets. Even though RHE performs better in MRR in most datasets, one can see MRR for RDHE is much better in congress-bills datasets. Hence one can conclude that using \texttt{Interaction update} stage resulted in performance improvement.

\paragraph{\textit{Effect of temporal drift on performance.}} Similar to the earlier study, the advantage of this stage can be observed by comparing the performance of RHE to DHE-drift and RDHE to DHE. There is an average of $44\%$ gain in MRR and $42.03\%$ reduction in MAE for DHE-drift compared to RHE. Further, DHE uniformly outperforms RDHE for interaction type prediction in all datasets. There is an average of $90\%$ improvement in the MRR for the interaction type prediction task and a $13.7\%$ decrease in the MAE error for the duration prediction task. Hence, \texttt{Temporal drift} stage helps in performance improvement.

\paragraph{\textit{Effect of history aggregation on performance.}} The advantage of this stage can be observed by comparing HGDHE to DHE and HGDHE-hist to DHE-drift. We can see HGDHE outperforms DHE in all the tasks except in Congress bills. There is an average gain of $1.7\%$ in MRR and a $31.75\%$ reduction in MAE for HGDHE compared to DHE. We can see HGDHE-hist outperforms DHE-drift in all settings. A similar observation can be made in bipartite datasets when comparing models HGBDHE and BDHE. Both models give a comparable performance for interaction duration prediction except for CastGenre dataset. For interaction type prediction HGBDHE achieves a gain of $7\%$ in MRR when compared to BDHE. 

\subsection{Hyperedge Size ($k$) Vs Performance} \label{sec:visualizations}

\begin{figure}[t]
	\centering
	\subfloat[][]{\includegraphics[width=0.21\textwidth]{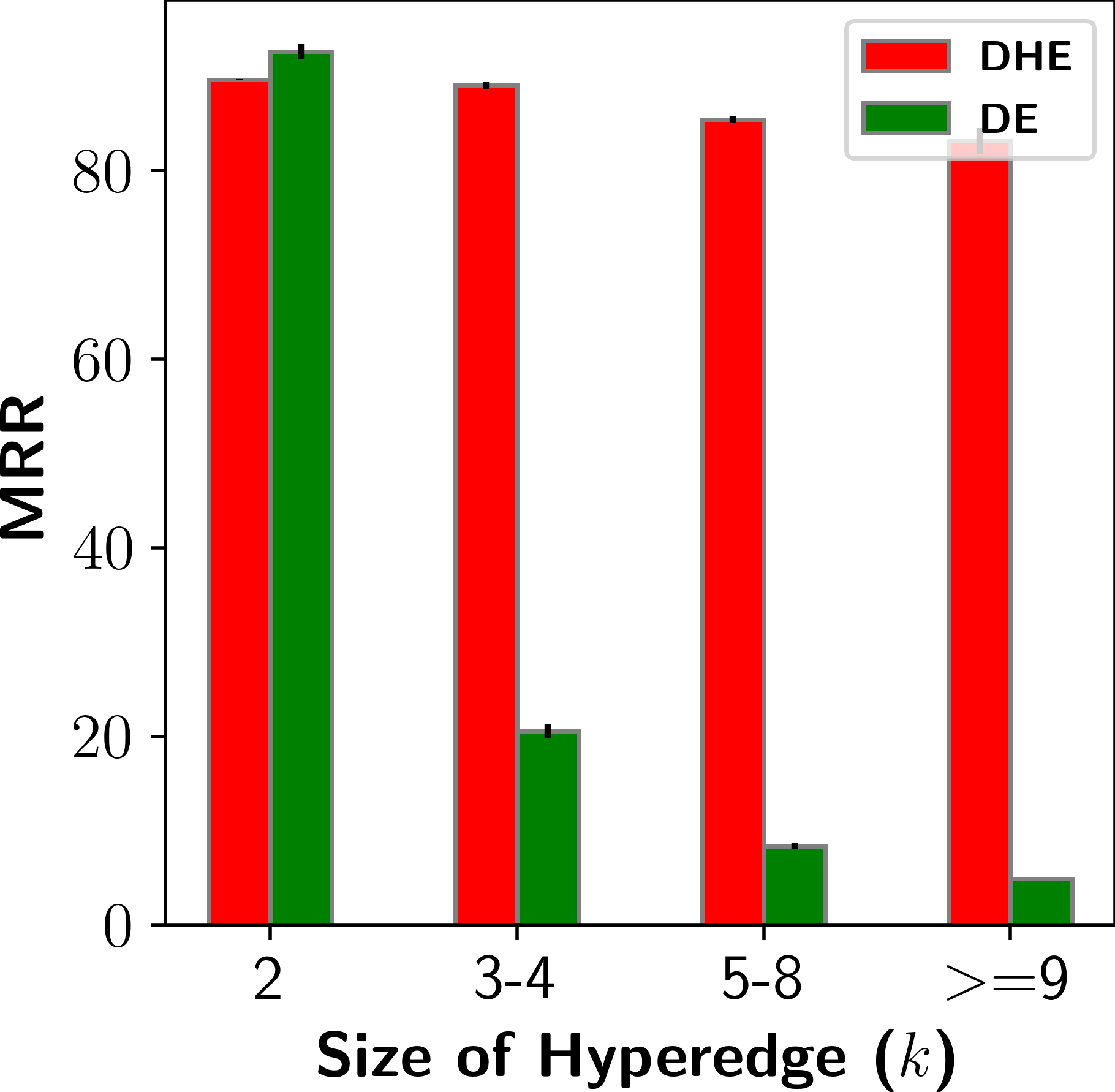}\label{fig:ablation_MRR_K}}%
	\subfloat[][]{ \includegraphics[width=0.21\textwidth]{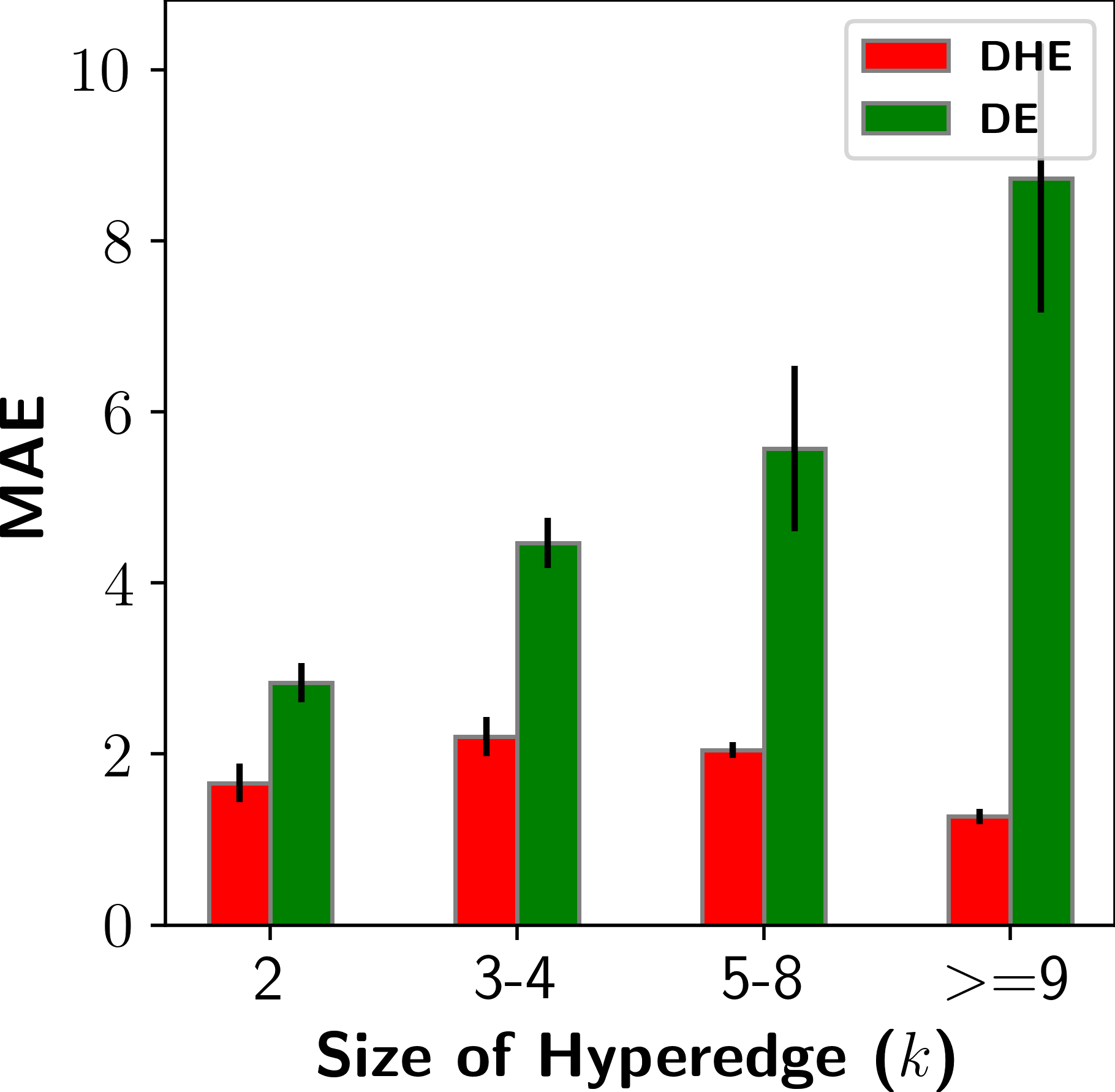}\label{fig:ablation_MAE_K}}%
	\caption{Figure~\ref{fig:ablation_MRR_K} shows the effect of hyperedge size ($k$) on interaction type prediction, and Figure~\ref{fig:ablation_MAE_K} shows the effect on interaction duration prediction for NDC-classes dataset}
\end{figure}

While both the models DHE and DE use dynamic representations, they have different conditional intensity functions. While DHE 
uses a hyperedge-based conditional intensity function, and DE uses a pairwise edge-based function. To compare them, we divide hyperedges into different groups based on their sizes, and the mean value of the evaluation metric is calculated for each group. These groups are defined so that each group has enough samples to make the comparisons statistically significant. In Figures~\ref{fig:ablation_MRR_K} and~\ref{fig:ablation_MAE_K}, we have shown the effect of hyperedge size on our model DHE, which uses hyperedge modeling, and compared it against model DE, which uses pairwise edges. Here, each hyperedge is grouped into groups with $k=2$, $3\leq k \leq 4$, $5 \leq k \leq 8$ and $k\geq9$. From Figure~\ref{fig:ablation_MRR_K}, we can see that performance of DE is poor for hyperedges with a size of more than 2, but our model DHE has almost similar performance for hyperedges of different sizes. The reason for this is that DHE is suited more for hyperedges of varying sizes compared to DE.  We observed a similar trend in other datasets, as shown in Appendix~\ref{sec:effect_K_on_performance}. For the interaction duration prediction task, the error for the DE model increases with hyperedge size while the DHE model performs similarly for all hyperedge sizes.


\section{Concluding Remarks}
\label{sec:conclucsion_and_future_work}
The problem of forecasting higher-order interactions in time-evolving hypergraphs is  very challenging, and this has not been studied before. In this work, we provide the first solution to this problem by proposing a model for forecasting higher-order interaction between nodes in a network as temporal hyperedge formation events. For this, we develop a mechanism that uses the temporal point process to learn the dynamic representation of nodes. Our future work includes extending the proposed model for modeling multi-relational higher orders interactions~\citep{Fatemi:EtAL:2020:Knowledge_Hypergraphs:_Prediction_Beyond_Binary_Relations} and incorporating multi-hop information into node representation using hypergraph neural network-based techniques for better predictive performance. We also plan to explore options 
to reduce the training time by using $t$-batch~\citep{Kumar:EtAL:2019:Predicting_Dynamic_Embedding_Trajectory_in_Temporal_Interaction_Networks} based approach that can use parallel training techniques.

\section*{Acknowledgements}
The authors would like to thank the SERB, Department of Science and Technology, Government of India, for the generous funding towards this work through the IMPRINT Project: IMP/2019/000383.
\bibliography{aaai23}

\newpage 

\appendix

\section{Notations}
\begin{table*}[!htbp]
	\centering
	\begin{tabular}{|c|c|}
		\toprule
		\textbf{Notation}  &  \textbf{Definition}  \\
		\midrule 
		$\lambda(t)$ & Conditional intensity function\\
		$t$ & Time \\
		$k$ & Hyperedge size \\ 
		$t^p$ & Previous time \\
		$\ud t$ & Derivative of time $t$ \\
		$\mathbb{R}$ & Real space with dimension 1 \\
		$\mathcal{V}$ & Set of nodes \\
		$\mathcal{T}$ & Historical event times \\
		$d$ & Embedding size \\
		$v_i$ & $i$th node \\
		$h_i$ & $i$th hyperedge/higher-order interaction \\
		$\mathcal{H}$ & Set of all valid interactions \\
		$\mathcal{E}$ & Sequence of temporal ordered interaction events \\
		$\lambda_h(t)$ & Conditional intensity function of interaction $h$\\
		$t^p_h$ & Previous time of interaction for hyperedge $h$ \\
		$t^s$ &  Monte-carlo $t$ sample \\
		$p(t)$/$p_h(t)$ & Probability of the event/interaction $h$ \\
		$S(t)$ & Survival function for the event \\ 
		$t^p_v $ & Previous time of interaction for node $v$ \\
		$t^l_h$ & Last time occurrence of $h$ \\
		$f(\  )$ & Scoring function for conditional intensity \\
		$\bm{\theta}$ & Learnable parameter \\
		$\bm{v}(t) $ & Node embedding at time $t$ \\
		$\bm{W_0}, \bm{W_1}, \bm{W_2}, \bm{b_0} $ & Learnable parameters for dynamic node embeddings \\
		$\bm{W_3}, \bm{W_4}, \bm{W_5}, \bm{b_1} $ & Learnable parameters for interaction update\\
		$\bm{v^s}(t) $ & History aggregation embeddings \\
		$\bm{\Phi}(t)$ & Time embedding function\\
		$\{\omega_i \}_{i=1}^d$ and  $\{\theta_i \}_{i=1}^d$  & Learnable parameters of $\Phi(t)$  \\
		$\bm{W_Q}, \bm{W_K}, \bm{W_V}$ & Self-Attention learnable weights\\
		$e_{ij}$ & Attention inner-product score \\
		$\alpha_{ij}$ & Attention weights \\
		$\bm{W_s}$ & Static embeddings linear layer \\
		$\bm{{d}_i^{h}}$ & Dynamic embeddings \\
		$\bm{{s}_i^{h}}$ & Static embeddings \\
		$o_i$ & Output Hadamard layer \\
		$\bm{W_o},b_o$ & Output linear layer \\
		$\mathcal{P}$ & Scoring function\\
		$\mathcal{K}()$ & Kernel function \\
		$\mu$ & Baseline intensity \\
		$\alpha $ & Weight parameter \\
		$\mathcal{B}$ & Negative sample size \\ 
		$N$ & Sampling for Monte-Carlo integration \\
		$M$ & Mini batch segment size \\ 
		\bottomrule
	\end{tabular}
	\caption{Notations}
	\label{tab: Notations}
\end{table*}
Table~\ref{tab: Notations} shows the notations and their respective definitions used in this paper. We used bold lower-case letters to show vectors and bold upper-case letters to show matrices.
\section{Background on Temporal Point Process} \label{sec:Temporal_Point_Process}
A TPP~\citep{Daley:EtAL:An_introduction_to_the_theory_of_point_processes.} is a continuous-time stochastic process that models discrete events in time ${t_i}$, $t_i \in \mathbb{R}^{+}$. It defines a conditional density function (CDF) for future event time $t$ by observing historical event times till time $t_n$, $\mathcal{T}(t_n) = \{t_1, t_2,..., t_n \}$. A convenient and interpretable way to parameterize CDF is by defining a conditional intensity function or instantaneous stochastic rate of events $\lambda(t)$ defined below, 
\begin{align} \label{eq: conditional intensity function}
	\lambda(t) \ud t = p( \text{event in} [t, t+ \ud t] |  \mathcal{T}(t) ) . 
\end{align}
That is, $\lambda(t) \ud t$ is the probability of observing an event in interval $[t, t+ \ud t]$ by observing history till $t$. Then we can write the CDF of the next event time as,
\begin{align} \label{eq: event probability}
	p(t|  \mathcal{T}(t_n) ) & = \lambda(t) \mathcal{S}(t) \nonumber , \\
	\mathcal{S}(t) &  =  \exp{ ( \int_{t_n}^t -\lambda (\tau) \ud \tau ) } .
\end{align}
Here,  $\mathcal{S}(t)$ is the probability that no event happened during the interval $[t_n^, t)$, which is called the survival function. 

The choice of the functional form of $\lambda(t)$ depends upon the nature of the problem we are trying to model. For example, the Poisson process has constant intensity $\lambda(t)  = \mu$. Hawkes process \citep{Alan:Spectra_of_Some_Self-Exciting_and_Mutually_Exciting_Point_Processes} is used when events have self exciting nature with $\lambda(t) = \mu + \alpha \sum_{t_i \in \mathcal{T}(t)} \mathcal{K}(t -t_i) $. Here, $\mu \geq 0$ is the base rate, $\mathcal{K}(t) \geq 0 $ is the excitation kernel, and $\alpha \geq 0 $ is the strength of excitation. Rayleigh process \citep{Ghosh:EtAL:2009:Survival_and_Event_History_Analysis} with $\lambda(t) = \alpha t$, where rate of events increase with time and $\alpha > 0$ is the weight parameter. If the intensity is the output of neural network $f( . )$ that takes history as input, $\lambda(t)= f(\mathcal{T}(t))$, it is called Neural Temporal Point process \citep{shchur:EtAL:2021:Neural_Temporal_Point_Processes}.


\section{Conditional Intensity Function Architecture} \label{sec:Conditional_Intensity_Function_Architecture}

\subsection{Homogeneous} \label{sec:lambda_homogeneous}
Given the node embeddings of hyperedge $h=\{v_1, v_2, \ldots, v_k\}$, the importance weights for each node are calculated by a self-attention layer, as given by, 
\begin{align}
	e_{ij} &= ( \bm{W_Q}^{T} \bm{v_i} (t))^T \bm{W_K}^{T} \bm{v_j} (t), \text{$\forall  1 \leq i,j \leq k, i \neq j $}, \nonumber \\
	\alpha_{ij} &= \frac{\displaystyle \exp{ (e_{ij}}) }{\displaystyle \sum_{ 1 \leq \ell \leq k, i \neq \ell } \exp{ (e_{i\ell}}) }.
\end{align}

These weights are used to calculate the dynamic embeddings for each node $v_i$ as, 
\begin{align} \label{eq:dynamic_embeddings_homogeneous}
	\bm{{d}_i^h}  = \tanh\left( \sum_{1 \leq j \leq k, i \neq j } \alpha_{ij} \bm{W_V}^{T} \bm{v_j} (t) \right) .
\end{align}
Here, $\bm{W_Q}, \bm{W_K}, \bm{W_V} \in \mathbb{R}^{d \times d}$ are learnable weights. We also create static embeddings $\bm{s_i^h} = \bm{W_s} \bm{v_i}(t)$, where $\bm{W_s} \in \mathbb{R}^{d \times d}$ is a learnable parameter. Then we calculate the Hadamard power of the difference between static and dynamic embedding pairs followed by a linear layer and average pooling to get the final score $\mathcal{P}^h$ as shown below,
\begin{align}
	o_i  &= \bm{W_o}^T ( \bm{{d}_i^h} - \bm{s_i^h}  ) ^ 2 + b_o, \nonumber \\ 
	\mathcal{P}^h   &=  \frac{1}{k}\sum_{i=1}^k \mathcal{P}_i^h  =\frac{1}{k}\sum_{i=1}^k \log( 1 + \exp{( o_i  ) } ) .
\end{align}
Here, $\bm{W_o} \in \mathbb{R}^{ d \times 1}, b_o \in \mathbb{R}$ are learnable parameters of output layer. In our model the value of $f( \bm{v_1} (t) , \bm{v_2} (t), \ldots, \bm{v_k} (t) )= \mathcal{P}^h$. 

\subsection{Bipartite}  \label{sec:lambda_bipartite}
For a bipartite hyperedge $h$, there are two sets of nodes with a directed relation between them. This can represented as  $h=(\{v_1, v_2, \ldots, v_k\}, \{v_{1^{'}}, v_{2^{'}}, \ldots, v_{k^{'}}\}$ ). Here, we use the cross attention between the sets of nodes to create dynamic embeddings, as shown below,

\begin{align}
	e_{ij^{'}}  &= ( \bm{W_Q}^{T} \bm{v_i} (t))^T \bm{W_{K^{'}}}^{T} \bm{v_{j^{'}}} (t), \text{$\forall  1\leq i \leq k, 1 \leq  j^{'} \leq k^{'} $}, \nonumber \\ 
	e_{i^{'}j}^{'}  &= ( \bm{W_{Q^{'}}}^{T} \bm{v_{i^{'}}} (t))^T \bm{W_K} \bm{v_j}^{T} (t), \text{$\forall  1\leq i^{'} \leq k^{'}, 1 \leq  j \leq k $} \nonumber \\
	\alpha_{ij^{'}} &= \frac{\displaystyle \exp{ (e_{ij^{'}}}) }{\displaystyle \sum_{ 1 \leq \ell^{'} \leq k^{'} } \exp{ (e_{i\ell^{'}}}) } \nonumber \\ 
	\alpha_{i^{'}j} &= \frac{\displaystyle \exp{ (e_{i^{'}j}}) }{\displaystyle \sum_{ 1 \leq \ell \leq k } \exp{ (e_{i^{'}\ell}}) }.
\end{align}
These weights are used to find the dynamic embeddings for left node $v_i$ and right node $v_{i^{'}}$ as follows, 

\begin{align} \label{eq:bipartite_dynamic_embeddings}
	\bm{{d}_i^h}  &= \tanh\left( \sum_{1 \leq j^{'} \leq k^{'}} \alpha_{ij^{'}} \bm{W_{V^{'}}}^{T} \bm{v_{j^{'}}}(t) \right) \nonumber \\ 
	\bm{ {d}_{i^{'}}^h }  &= \tanh\left( \sum_{1 \leq j \leq k} \alpha_{i^{'}j} \bm{W_{V}^{T} v_{j} }(t) \right) .
\end{align}

Here, $ \bm{W_Q}, \bm{W_K}, \bm{W_V} \in \mathbb{R}^{d \times d} $ are the learnable weights for nodes in the left, and $ \bm{W_{Q^{'}}}, \bm{W_{K^{'}}}, \bm{W_{V^{'}}} \in \mathbb{R}^{d \times d} $ are the learnable weights for nodes in the right. Then static embeddings are created for both left $\bm{s_i^h} = \bm{W_s} \bm{v_i}(t)$ and right side $\bm{{s}_{i^{'}}^h} = \bm{W_{s^{'}}} \bm{v_{i^{'}}} (t)$, where $\bm{W_s}, \bm{W_{s^{'}}} \in \mathbb{R}^{d \times d}$ are learnable parameters. Then we calculate the Hadamard power of the difference between static and dynamic embedding pairs for both the left and right nodes, followed by a linear layer and average pooling to get final score $\mathcal{P}^h$ as shown below,

\begin{align}
	o_i  &= \bm{W_o}^T ( \bm{{d}_i^h} - \bm{s_i^h}  ) ^ 2+ b_o, \nonumber \\
	o_{i^{'}}  &= \bm{W_{o^{'}}}^T ( \bm{ {d}_{i^{'}}^h } - \bm{{s}_{i^{'}}^h} ) ^ 2 + b_{o^{'}}, \nonumber \\
	\mathcal{P}^h   &=  \frac{1}{k}\sum_{i=1}^k \log( 1 + \exp{( o_i  ) } ) + \frac{1}{k^{'}}\sum_{i=1}^k \log( 1 + \exp{( o_{i^{'}}  ) } ).
\end{align}

Here, $\bm{W_o}, \bm{W_{o^{'}}}  \in \mathbb{R}^{ d \times 1}$ and $b_{o^{'}}, b_o \in \mathbb{R}$ are learnable parameters of output layer. In our model the value of $f( \bm{v_1} (t) , \bm{v_2} (t), \ldots, \bm{v_k} (t) )= \mathcal{P}^h$. 
\section{Datasets}
\label{sec:datasets}

\paragraph{Email Network} [email-Enron \citep{klimt:EtAL:2004:Enron}; email-Eu \citep{Paranjape:EtAL:2017:Motifs_in_temporal_networks}]. This is a sequence of timestamped email interactions between employees at a company. The entities involved are the email addresses of the sender and receivers.  The timestamps in this are recorded at a resolution of 1-millisecond, which are scaled down in all our experiments.  The scaling factor is chosen as the median value of interevent duration for both datasets.


\paragraph{congress-bills} \citep{Fowler:EtAL:2006:Legislative}. Here, interactions are the legislative bills put forth in the House of Representatives and the Senate in the US. The entities involved in an interaction are the US congresspersons who sponsor and co-sponsor the bill. The timestamp in this is the date when the bill is introduced.

\paragraph{Drug Networks [NDC-classes, NDC-sub].}  Each interaction is a sequence of drugs, and timestamps are the dates at which the drugs were introduced in the market. In NDC classes, nodes involved in the interaction are class labels applied to the drugs. In NDC-sub, nodes involved in the interaction are substances that make up the drug.

\paragraph{[CastGenre, CastKeyword, CastCrew]} These datasets are prepared from the subset of Kaggle's The Movie Dataset~\footnote{https://www.kaggle.com/datasets/rounakbanik/the-movies-dataset}. For the CastGenre dataset, nodes in the left hyperedge are movie actors and the genres associated with the movie in the right hyperedge. For the CastKeyword dataset, we use keywords associated with movie plots for right hyperedge instead of genres. Here, the movie release date is the timestamp associated with each hyperedge. For the CastCrew dataset, we use the name of the crew associated with movie production for the right hyperedge.  We only considered movies released after 1990 and followed the same preprocessing as the recent work \citep{Sharma:2021:CATSET} for preparing these datasets.

\section{Baselines}
\label{appendix:baselines}

\paragraph{Rayleigh Hyperedge (RHE).} In this, we keep embeddings $\bm{v}(t)$ fixed for every time instance. Then intensity is modeled as the Rayleigh process as in Equation~\ref{eq:rayleighintensity}, and duration predictions are made using the closed form expression in Equation~\ref{eq:durationpredictionrayliegh}. 

\paragraph{Rayleigh Dynamic Hyperedge (RDHE).} Similar to RHE, but here we allow the node embeddings $\bm{v}(t)$ to evolve when an interaction involving the node $v$ occurs as per the \texttt{Interaction Update} stage but not through \texttt{Temporal Drift} and \texttt{History Aggregation} stage. So, $\bm{v}(t)$ is piece-wise continuous (node embeddings are constant during interevent time) as we do not allow the model to evolve the node embedding during the interevent time. This model is based on DeepCoevolve~\citep{Dai:EtAL:2016:Deep_coevolutionary_network:_Embedding_user_and_item_features_for_recommendation} model used for sequential recommendation. 

\paragraph{Dynamic Edge-Drift (DE-Drift).} In this model, each hyperedge event is modeled as a concurrent pairwise edge events. For example, $h = \{v_1, v_2, v_3 \}$ is modeled as concurrent edge events $\{ (v_1, v_2), (v_2, v_3), (v_1, v_3)\}$. Here, node embeddings $\bm{v}(t)$ are allowed to evolve during the interevent time using \texttt{Temporal Drift} stage. For predicting interaction at time $t$, we use the product of all conditional intensity functions of concurrent edge events (e.g. $\lambda_{v_1,v_2}(t),\lambda_{v_2,v_3}(t),\lambda_{v_1,v_3}(t) $ for interaction $h$). For predicting the time of interaction, we calculate $\hat{t}$ in Equation~\ref{eq:durationprediction} for each edge in the interaction and average them. In this model, conditional intensity is defined as $\lambda_{v_1 (t) , v_2 (t) } = \bm{v_1}^T (t) \bm{v_2} (t)$. 
\paragraph{Dynamic Edge (DE).} This uses the same modeling technique as DE-Drift. In addition, it also uses \texttt{Interaction Update} stage to evolve embeddings.

\paragraph{Dynamic Hyperedge-Drift (DHE-Drift).} This uses hyperedge modeling explained in Section~\ref{sec:DynamicNodeRepresentation}, but do not use \texttt{Interaction Update} and \texttt{History Aggregation} stage.

\paragraph{Dynamic Hyperedge (DHE).} This uses hyperedge modeling explained in Section~\ref{sec:DynamicNodeRepresentation}, but does not use \texttt{History Aggregation} stage.

\paragraph{Hypergraph Dynamic Hyperedge-hist (HGDHE-hist). } This uses hyperedge modeling explained in Section~\ref{sec:DynamicNodeRepresentation}, but does not use \texttt{Interaction Update} stage.


\paragraph{Bipartite Dynamic Edge (BDE).}   This is the bipartite version of DE, where each bi-partite hyperedge is modeled as concurrent bipartite edges. For example, $h = ( \{v_1, v_2 \}, \{ v_{1^{'}}, v_{2^{'}} \} )$ is modeled as concurrent edge events $\{ (v_1, v_{1^{'} }), (v_2, v_{1^{'}}),  (v_1, v_{2^{'}}), (v_2, v_{2^{'}})\}$. For learning node representation, the left and right nodes have a different set of parameters as explained in Section \ref{sec:bipartite_hyperedge_extension}. 

\paragraph{Bipartite Dynamic Hyperedge (BDHE). } This is the bipartite version of DHE. 

\paragraph{DyReP.}\cite{trivediEtAL:2019:DyRep:LearningRepresentationsoverDynamicGraphs} This is designed for forecasting pairwise links in a dynamic graph. The architecture of this is similar to \textbf{DE} model, however it uses graph neural network based neighborhood aggregation in the \texttt{Interaction Update Stage} and do not use \texttt{Temporal Drift} to evolve the embedding during the interevent time. We used our own implementation of this model to get comparable performance.

\paragraph{TGAT.} \cite{Xu:EtAL:2020:Inductive_representation_learning_on_temporal_graphs} This is a dynamic link prediction model for temporal graphs. For learning dynamic node embeddings, it uses graph attention to aggregate the topological features by considering the temporal history of the data. We transferred the hypergraph dataset to pairwise edge by clique expansion and used the code \footnote{https://github.com/StatsDLMathsRecomSys/Inductive-representation-learning-on-temporal-graphs} provided by the authors to learn the model. 

\paragraph{TGN.} \cite{Rossi:EtAL:2020:Temporal_graph_networks_for_deep_learning_on_dynamic_graphs} This model is for link prediction in temporal graphs. In this, interactions are stored in the form of messages and are used to update the respective nodes' memory modules. This memory embedding is used as an input to the temporal graph attention module for getting the dynamic node embedding. We followed the same preprocessing as for TGAT and used the code \footnote{https://github.com/twitter-research/tgn} provided by the authors to learn the model.



\section{Computational Infrastructure} 
\label{appendix:computational_infrastructure}
We used a workstation with AMD EPYC 7742 processor, Nvidia-DGX-A100 with 512 GB of main memory, and 4 x A100 GPUs with 40 GB of memory.
\section{Effect of Hyperedge Size on Performance}\label{sec:effect_K_on_performance}
\begin{figure}[t]
	\centering
	\subfloat[][]{\includegraphics[width=0.21\textwidth]{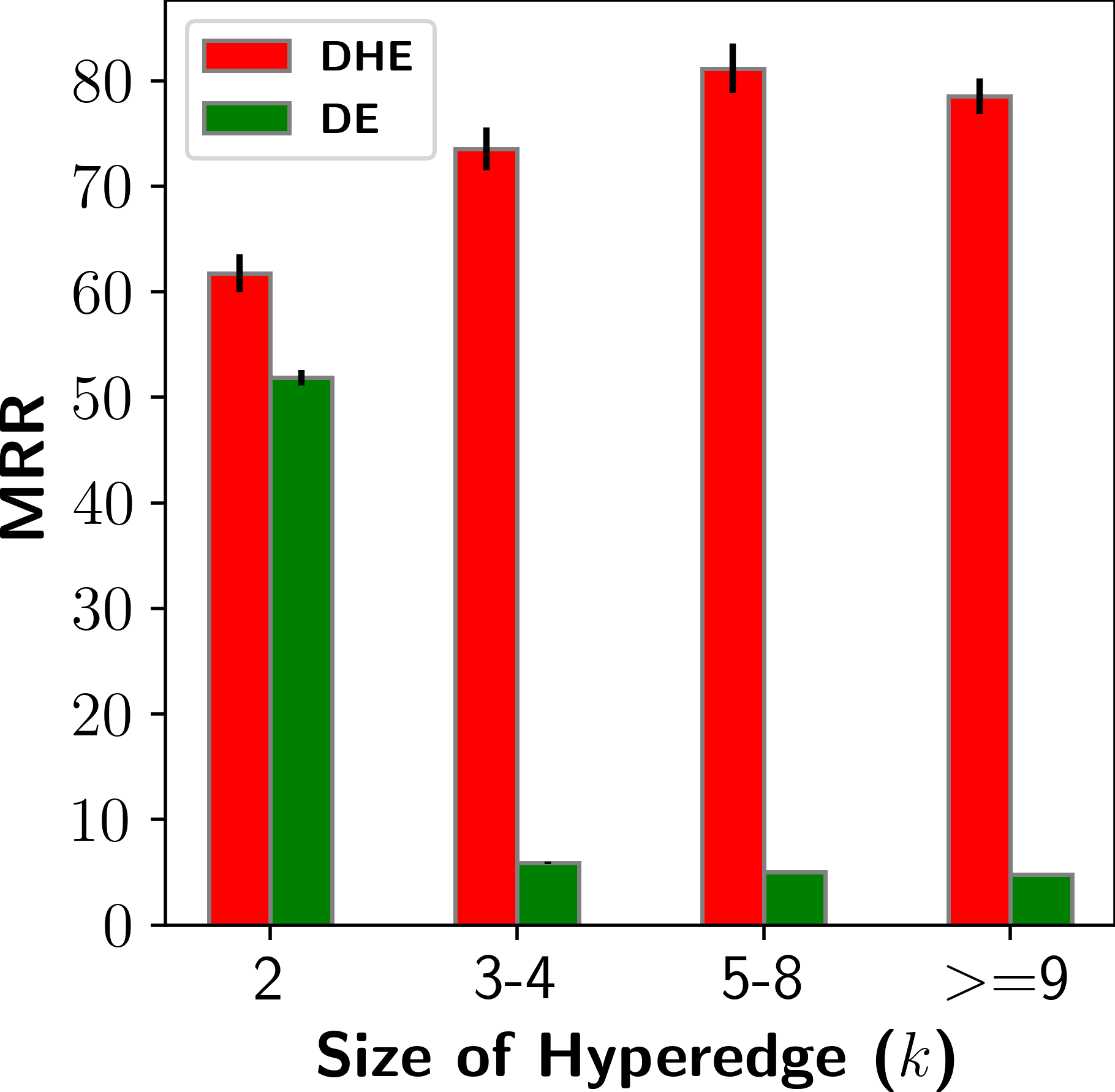}\label{fig:ablation_MRR_K_eu}}%
	\subfloat[][]{\includegraphics[width=0.20\textwidth]{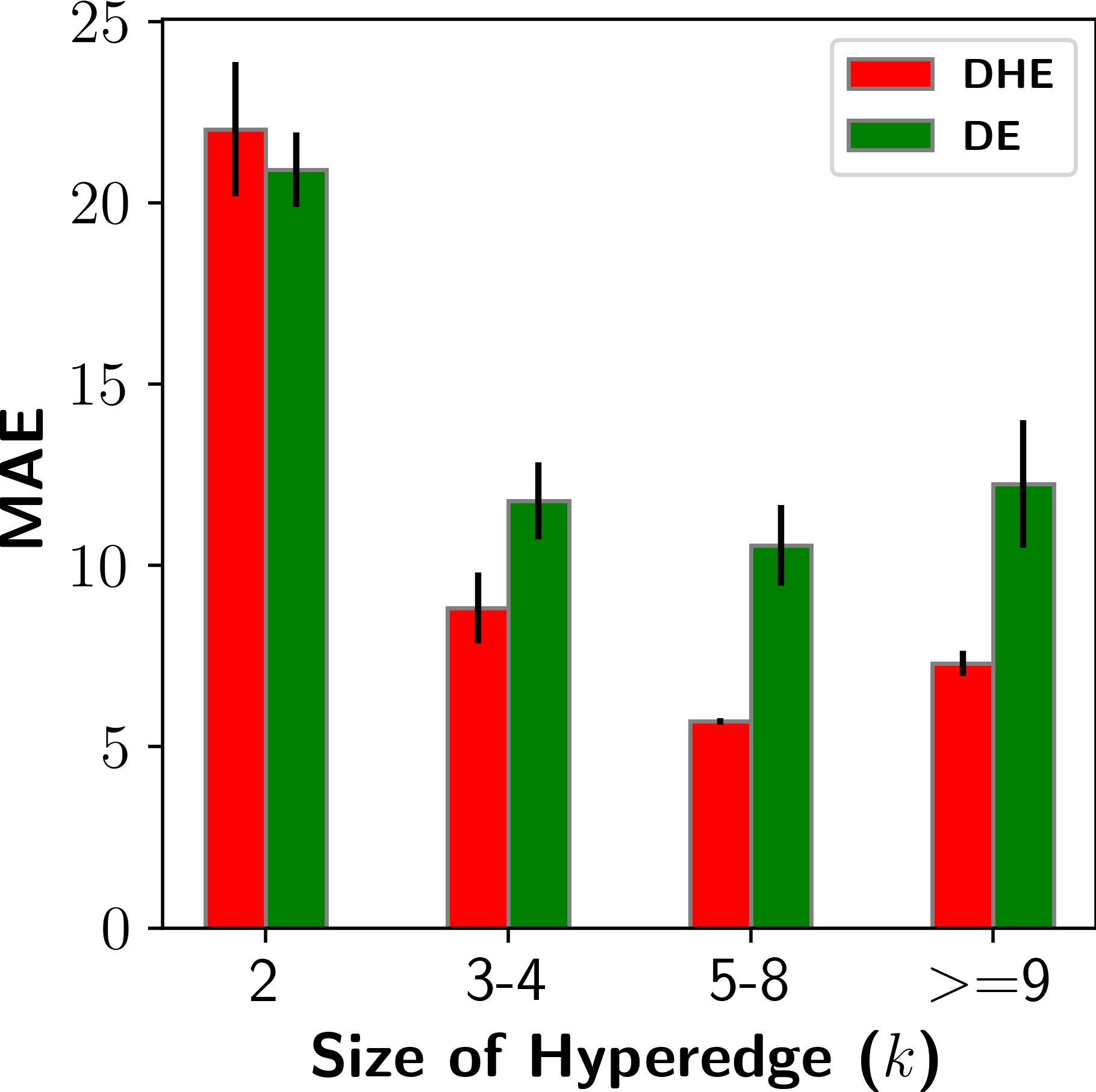}\label{fig:ablation_MAE_K_eu}}%
	\caption{Figure \ref{fig:ablation_MRR_K_eu} shows the effect of hyperedge size ($k$) on interaction type prediction, and Figure \ref{fig:ablation_MAE_K_eu} shows the effect on interaction duration prediction for email-Eu dataset}
	\label{fig:eumail}
\end{figure}

\begin{figure}[t]
	\centering
	\subfloat[][]{\includegraphics[width=0.21\textwidth]{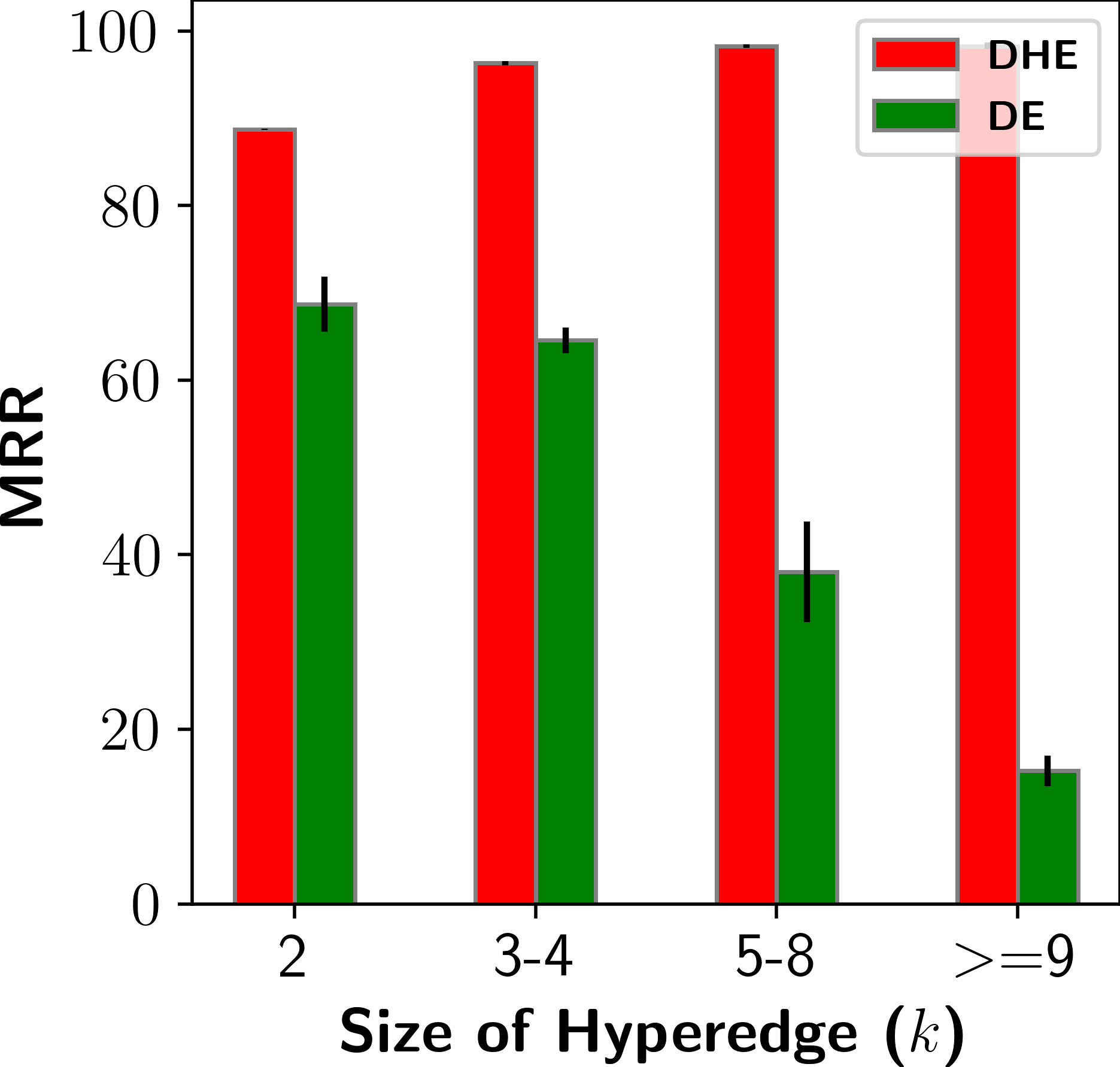}\label{fig:ablation_MRR_K_cong}}%
	\subfloat[][]{\includegraphics[width=0.21\textwidth]{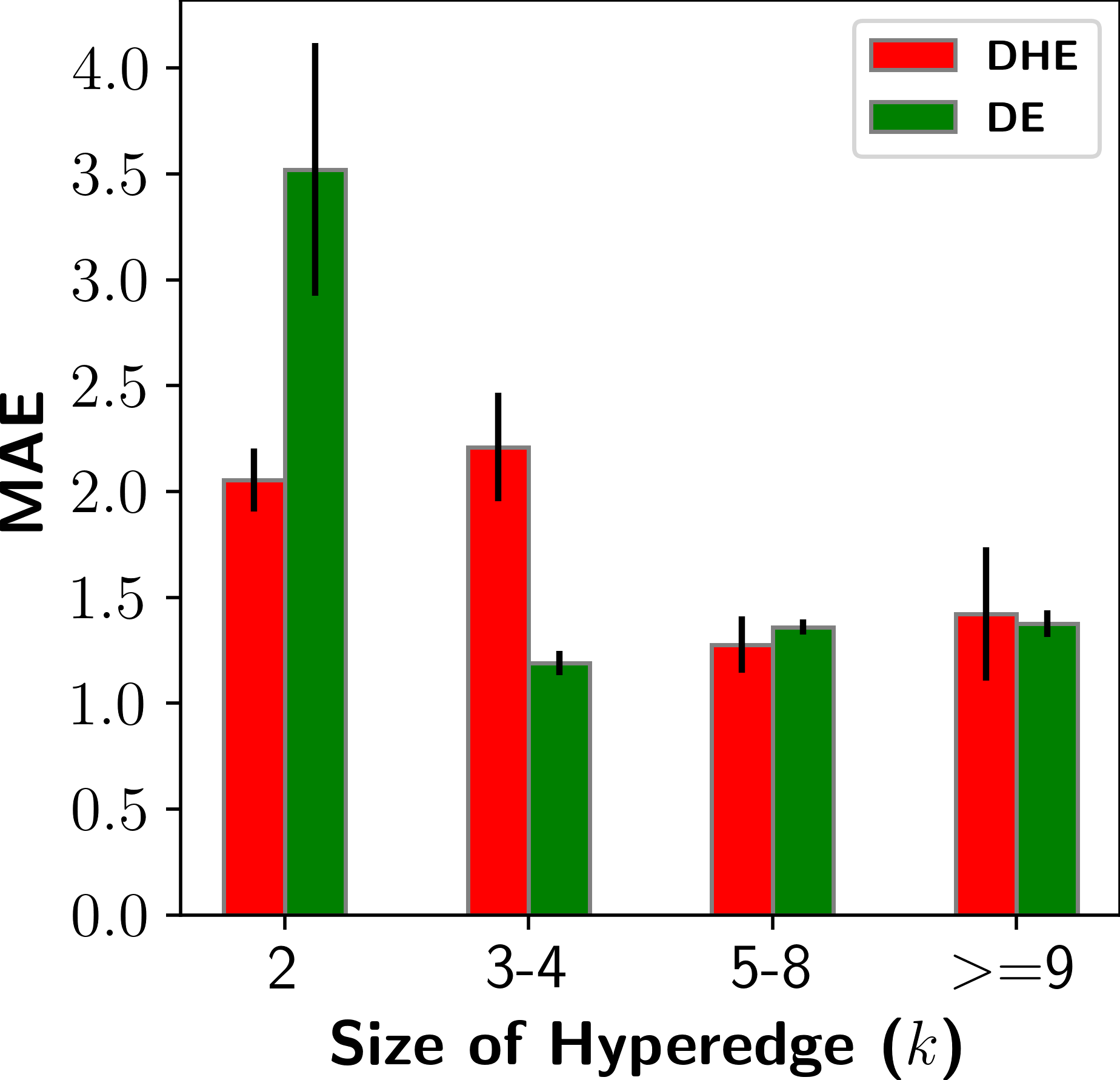}\label{fig:ablation_MAE_K_cong}}%
	\caption{Figure \ref{fig:ablation_MRR_K_cong} shows the effect of hyperedge size ($k$) on interaction type prediction, and Figure \ref{fig:ablation_MAE_K_cong} shows the effect on interaction duration prediction for congress-bills  dataset}
	\label{fig:congress}
\end{figure}

\begin{figure}[t]
	\centering
	\subfloat[][]{\includegraphics[width=0.21\textwidth]{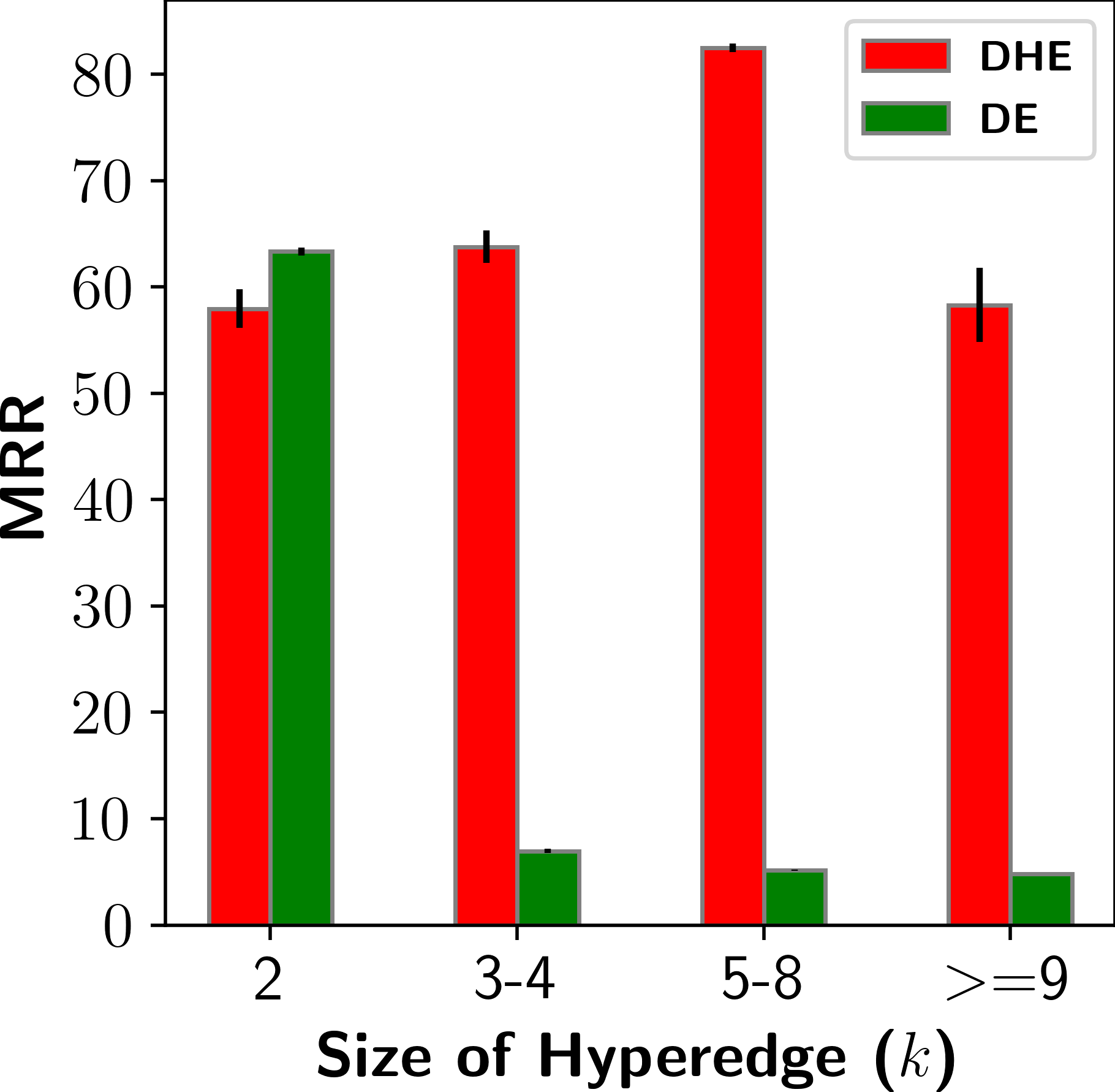}\label{fig:ablation_MRR_K_enron}}%
	\subfloat[][]{\includegraphics[width=0.21\textwidth]{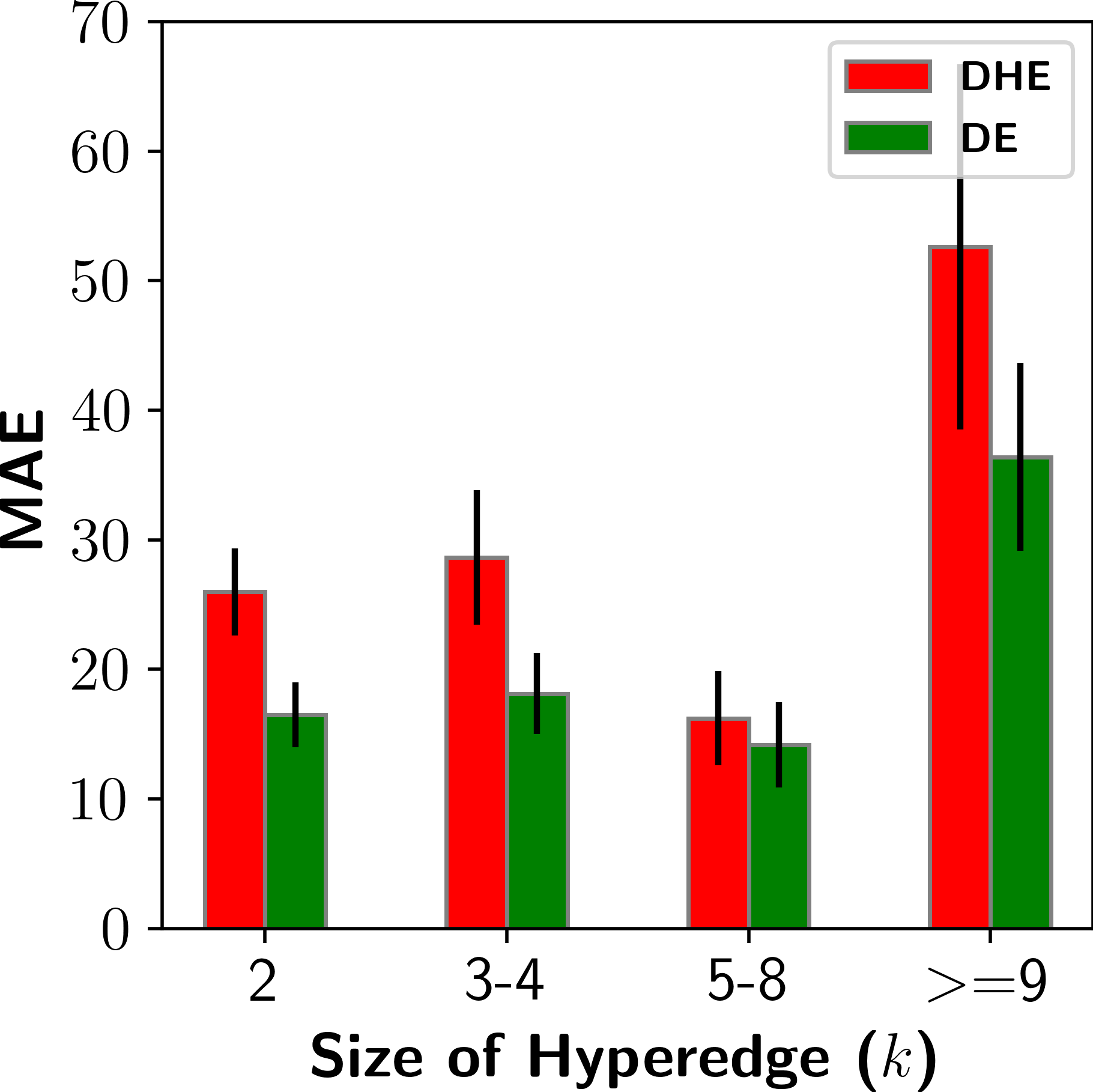}\label{fig:ablation_MAE_K_enron}}%
	\caption{Figure \ref{fig:ablation_MRR_K_enron} shows the effect of hyperedge size ($k$) on interaction type prediction, and Figure \ref{fig:ablation_MAE_K_enron} shows the effect on interaction duration prediction for Enron dataset}
	\label{fig:enron}
\end{figure}

\begin{figure}[t]
	\centering
	\subfloat[][]{\includegraphics[width=0.21\textwidth]{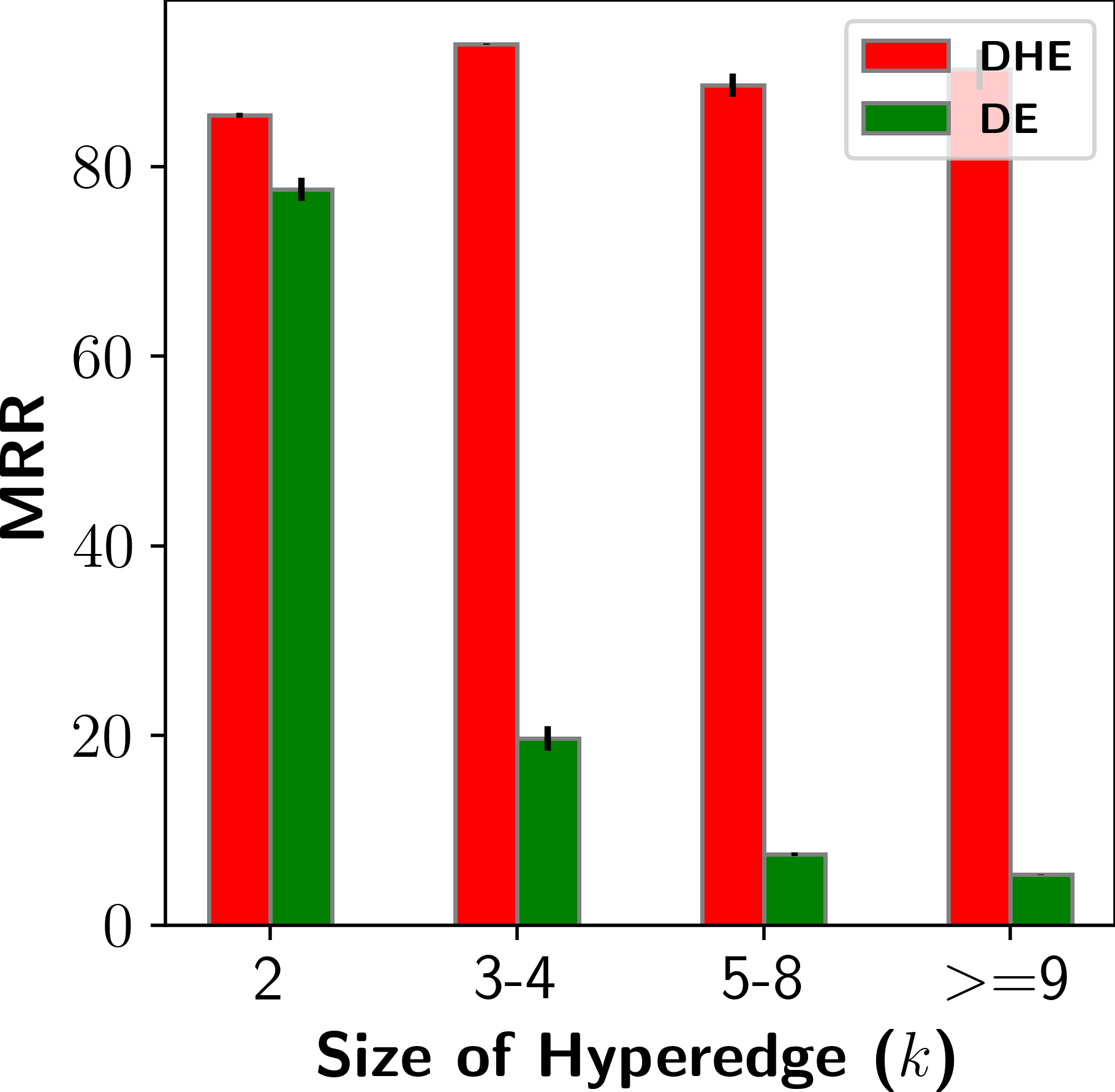}\label{fig:ablation_MRR_K_ndcsub}}%
	\subfloat[][]{\includegraphics[width=0.21\textwidth]{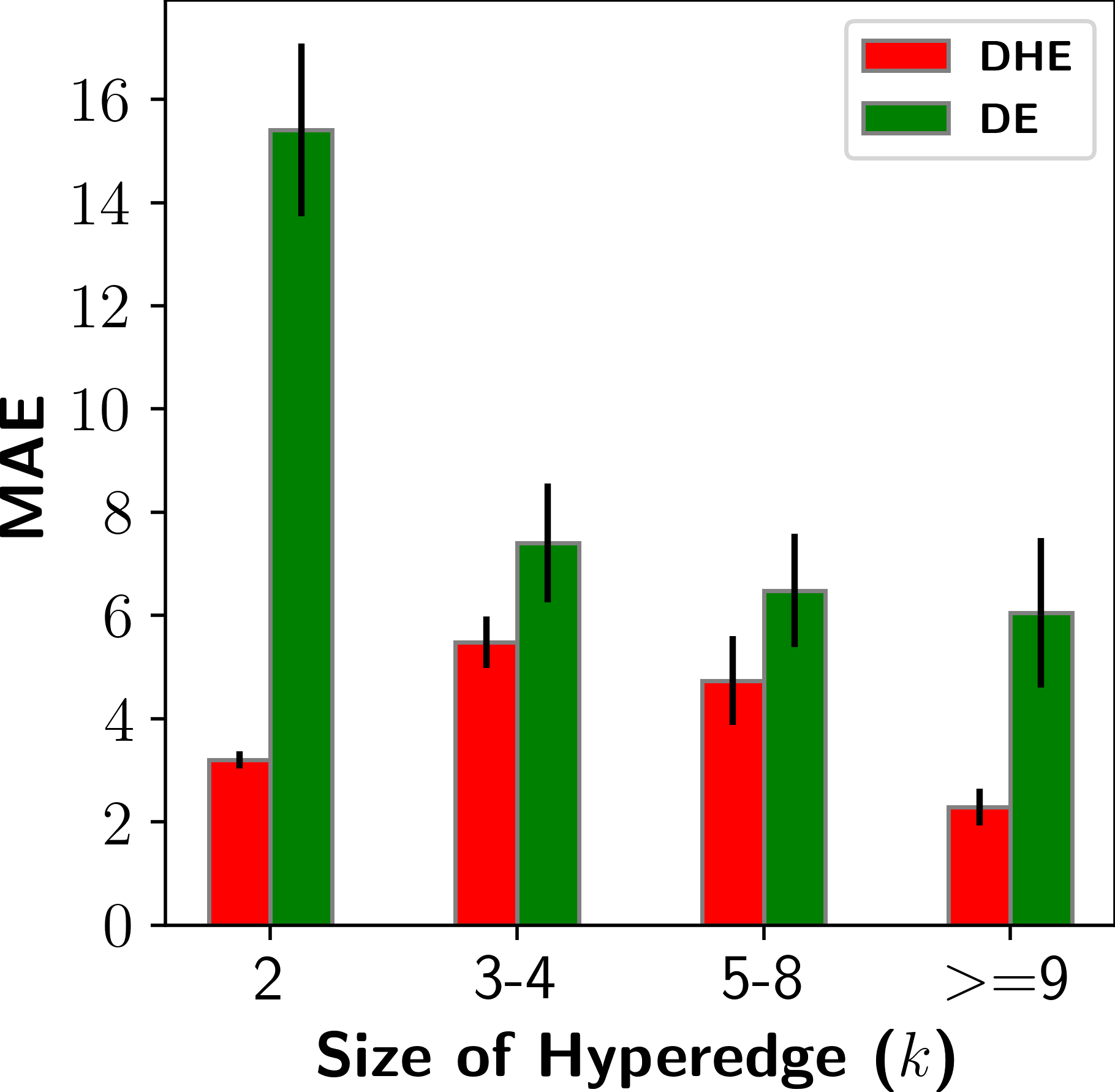}\label{fig:ablation_MAE_K_ndcsub}}%
	\caption{Figure \ref{fig:ablation_MRR_K_ndcsub} shows the effect of hyperedge size ($k$) on interaction type prediction, and Figure \ref{fig:ablation_MAE_K_ndcsub} shows the effect on interaction duration prediction for NDC-sub  dataset}
	\label{fig:ndcsub}
\end{figure}

The results for analysis on hyperedge size on performance explained in Section \ref{sec:visualizations} for datasets email-Eu, congress-bills, NDC-classes, and NDC-sub are shown in Figures \ref{fig:eumail}, \ref{fig:congress}, \ref{fig:enron}, and \ref{fig:ndcsub}, respectively. In all figures, we observe that MRR for interaction type prediction is poor for DE compared to DHE for hyperedge of size ($k$) more than 2.

\end{document}